\documentclass[final]{cvpr}

\usepackage{times}
\usepackage{epsfig}
\usepackage{graphicx}
\usepackage{amsmath}
\usepackage{amssymb}
\DeclareMathOperator*{\argmax}{arg\,max}
\usepackage[ruled,vlined]{algorithm2e}
\usepackage{algorithmic}
\usepackage{caption}
\usepackage{subcaption}
\usepackage[T1]{fontenc}
\pdfoutput=1

\usepackage[pagebackref=true,breaklinks=true,colorlinks,bookmarks=false]{hyperref}

\def\cvprPaperID{10725} 
\def\confYear{CVPR 2021}

\begin{document}

\title{A Singular Value Perspective on Model Robustness}

\author{Malhar Jere\\
UC San Diego\\
{\tt\small mjjere@ucsd.edu}
\and
Maghav Kumar\\
UIUC\\
{\tt\small mkumar10@illinois.edu}

\and
Farinaz Koushanfar\\
UC San Diego\\
{\tt\small fkoushanfar@ucsd.edu}}
\maketitle

\begin{abstract}
  
  Convolutional Neural Networks (CNNs) have made significant progress on several computer vision benchmarks, but are fraught with numerous non-human biases such as vulnerability to adversarial samples. Their lack of explainability makes identification and rectification of these biases difficult, and understanding their generalization behavior remains an open problem. In this work we explore the relationship between the generalization behavior of CNNs and the Singular Value Decomposition (SVD) of images. We show that naturally trained and adversarially robust CNNs exploit highly different features for the same dataset. We demonstrate that these features can be disentangled by SVD for ImageNet and CIFAR-10 trained networks. Finally, we propose \textbf{R}ank \textbf{I}ntegrated \textbf{G}radients \textbf{(RIG)}, the first rank-based feature attribution method to understand the dependence of CNNs on image rank. 
\end{abstract}

\vspace{-20pt}

\section{Introduction}

Deep Neural Networks have made significant progress on several challenging tasks in computer vision such as image classification ~\cite{imagenet_classification_first}, object detection~\cite{redmon2016you} and semantic segmentation~\cite{he2017mask, messaoud2020can}.
However, these networks have been shown to possess numerous non-human biases, such as high facial recognition misclassification error rates against certain races and genders~\cite{buolamwini2018gender}, vulnerability to numerous classes of adversarial samples~\cite{szegedy2013intriguing, goodfellow2014explaining, hosseini2018semantic, hendrycks2019natural, jere2019scratch, neekhara2020adversarial}, and vulnerability to training-time backdoor attacks ~\cite{liu2017trojaning}.

\begin{figure}[ht!]
    \centering
    \includegraphics[width=0.47\textwidth]{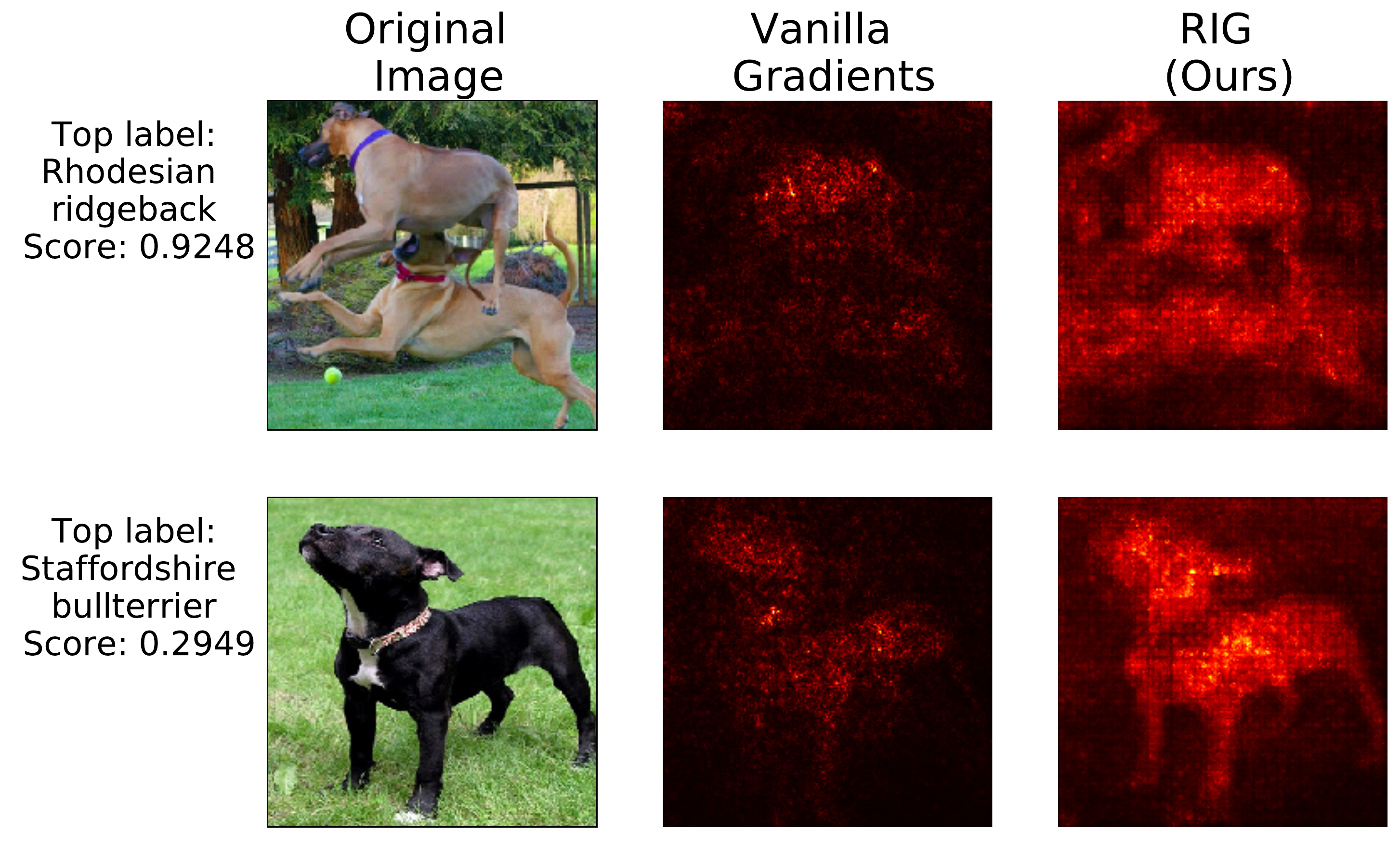}
    \caption{Left-to-right: The original image, Vanilla Gradients (as obtained by backpropagation with respect to the top label), and \textbf{R}ank \textbf{I}ntegrated \textbf{G}radients (RIG), our pixel importance method for the top label that averages saliency map information across low-rank representations of the same image. Notice that the visualizations obtained from RIG are better at identifying distinctive features of the image.}
    \label{fig:RIG}
\end{figure}

A line of recent efforts focused on explaining the generalization behavior of neural networks through adversarial robustness has shown significant promise. Such methods involve characterizing network inputs based on robust and non-robust features~\cite{ilyas2019adversarial, engstrom2019discussion, wang2020high}, understanding their effects on feature maps~\cite{xie2019feature}, interpreting their frequency components~\cite{fourier_perspective,fourier2} and interpreting their principal component properties~\cite{jere2019principal,bhagoji2017dimensionality}. 
Surprisingly, prior work has shown that neural nets often generalize to test sets based on superficial correlations in the training set~\cite{ilyas2019adversarial, wang2020high, geirhos2018imagenet, WangHLX19, Makino2020Differences}.

\begin{figure}
    \centering
    \includegraphics[width=0.45\textwidth]{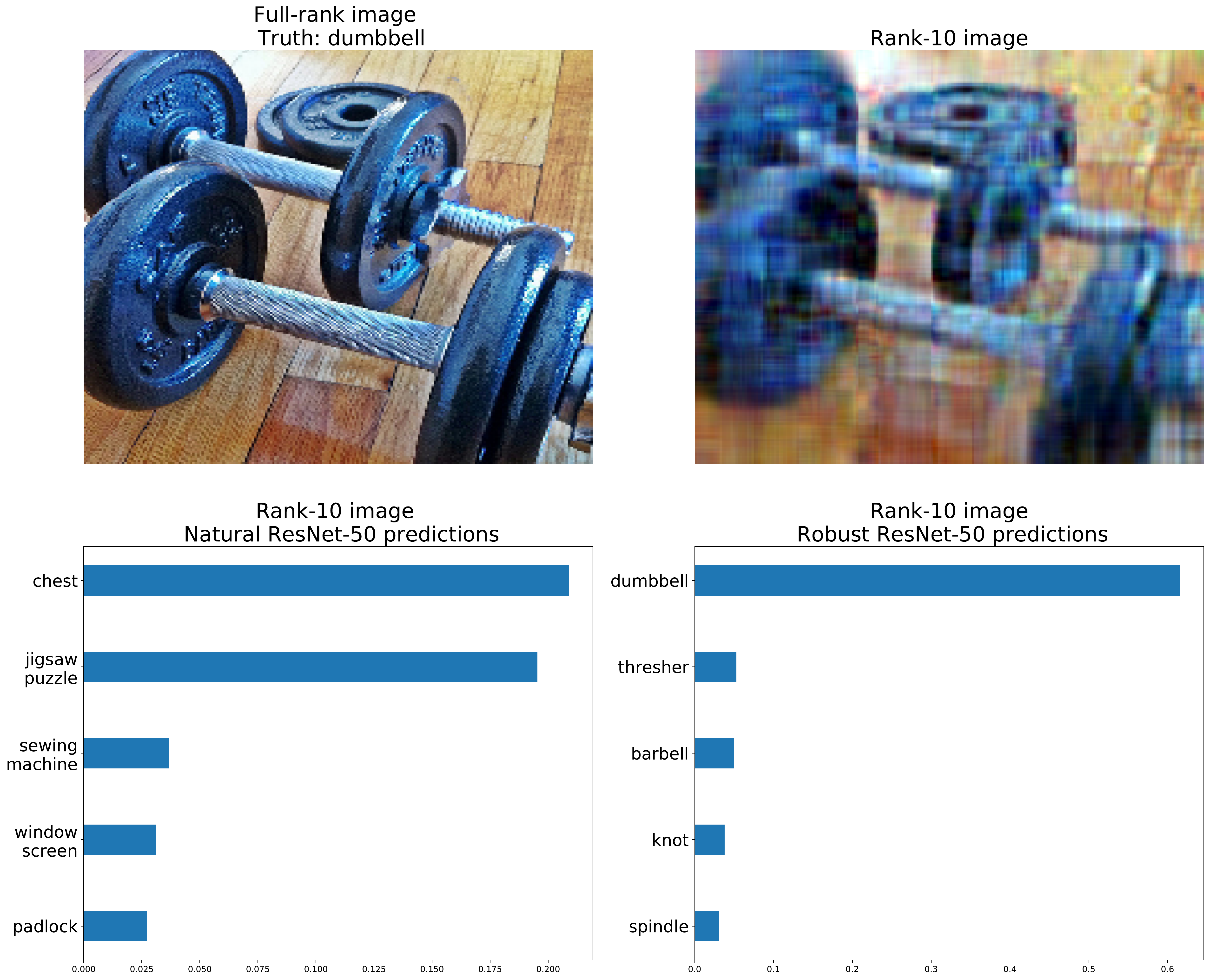}
    \caption{Accuracy for full-rank and rank-10 images. Top left is the full-rank image, top-right is its rank-10 approximation, bottom left are top-5 predictions for a naturally trained ResNet-50 on the rank-10 image, bottom right are top-5 predictions for an $L_{\infty}=4/255$ adversarially robust ResNet-50 on the rank-10 image. Note that the adversarially robust version makes better predictions on rank-based distorted versions of the same image, even though they do not fall under an $L_{p}$-distortion framework.}
    \label{fig:analogous_fig_1}
    \vspace{-0.4cm}
\end{figure}

In this work and inspired by previous works~\cite{jo2017measuring, ilyas2019adversarial, wang2020high}, we investigate the hypothesis that naturally trained CNNs leverage such superficial correlations in the dataset. However, different from prior works,
we argue that these superficial correlations can be distilled from an image via low-rank image approximation, a claim that was previously refuted~\cite{wang2020high}. We further argue that naturally trained neural networks and adversarially robust neural networks exploit highly different features from the same image, and that these features can be separated by singular value decomposition (SVD). Our contributions are as follows:

\begin{figure*}[ht!]
    \centering
    \includegraphics[width=0.75\textwidth]{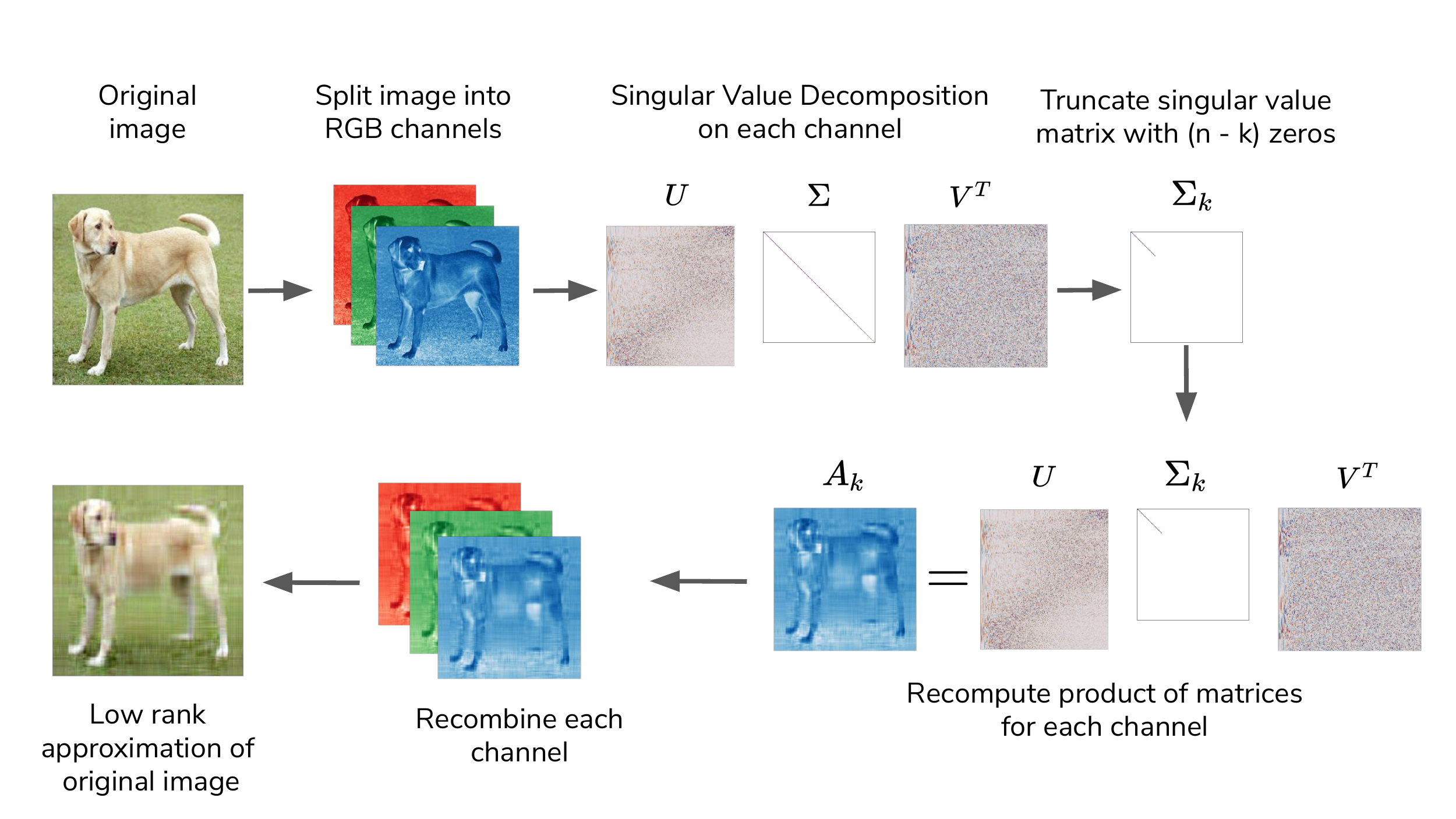}
    \caption{Generating a rank-$k$ image via truncated SVD. Given an $n \times n$ RGB image, we decompose the image into its individual color channels, zero out the last $(n - k)$ singular values obtained via SVD, and then reconstruct the image with its $k$ nonzero singular values. Low rank images are often more blurry than their full-rank counterparts.}
    \label{fig:low_rank_generation_methodology}
\end{figure*}
\begin{itemize}
    \item We identify for the first time that image rank (obtained from SVD) yields several novel insights about CNN robustness and interpretability (for example Figure~\ref{fig:analogous_fig_1}). We provide arguments in favor of using image rank as a potential human-aligned image robustness metric.

    \item We show empirically that naturally trained CNNs place a large importance on human-imperceptible higher-rank components, and that adversarial retraining increases reliance on human-aligned lower-ranked components. Furthermore, we demonstrate that neural networks trained on imperceptible, higher-rank features generalize to the test set.
    
    \item We provide experimental evidence that neural networks trained on low-rank images are more adversarially robust than their naturally trained counterparts for the same dataset, and capture the accuracy-robustness tradeoff in CNNs in this new lens.
    
    \item We propose \textbf{R}ank-\textbf{I}ntegrated \textbf{G}radients (RIG), the first rank-based feature attribution method. Saliency maps generated by RIG highlight features more in line with human vision and offer a new way to interpret the decisions of CNNs (Figure~\ref{fig:RIG}).
    
\end{itemize}

Our work provides a new methodology to capture model robustness, and allows us to distinguish between naturally trained and robust models outside of the traditional $L_{p}$-norm robustness framework. It suggests that data approximation strategies such as low-rank approximation can be leveraged to improve out-of-distribution CNN performance, such as against adversarial samples. Finally, we show that saliency maps that incorporate rank information highlight more visually meaningful features. We hope our work will encourage researchers to include image approximation techniques when studying CNN generalization.

\section{Background and Related Work} ~\label{sec2}
\vspace{-0.6cm}

\begin{figure*}[ht!]
    \centering
    \includegraphics[width=0.75\textwidth]{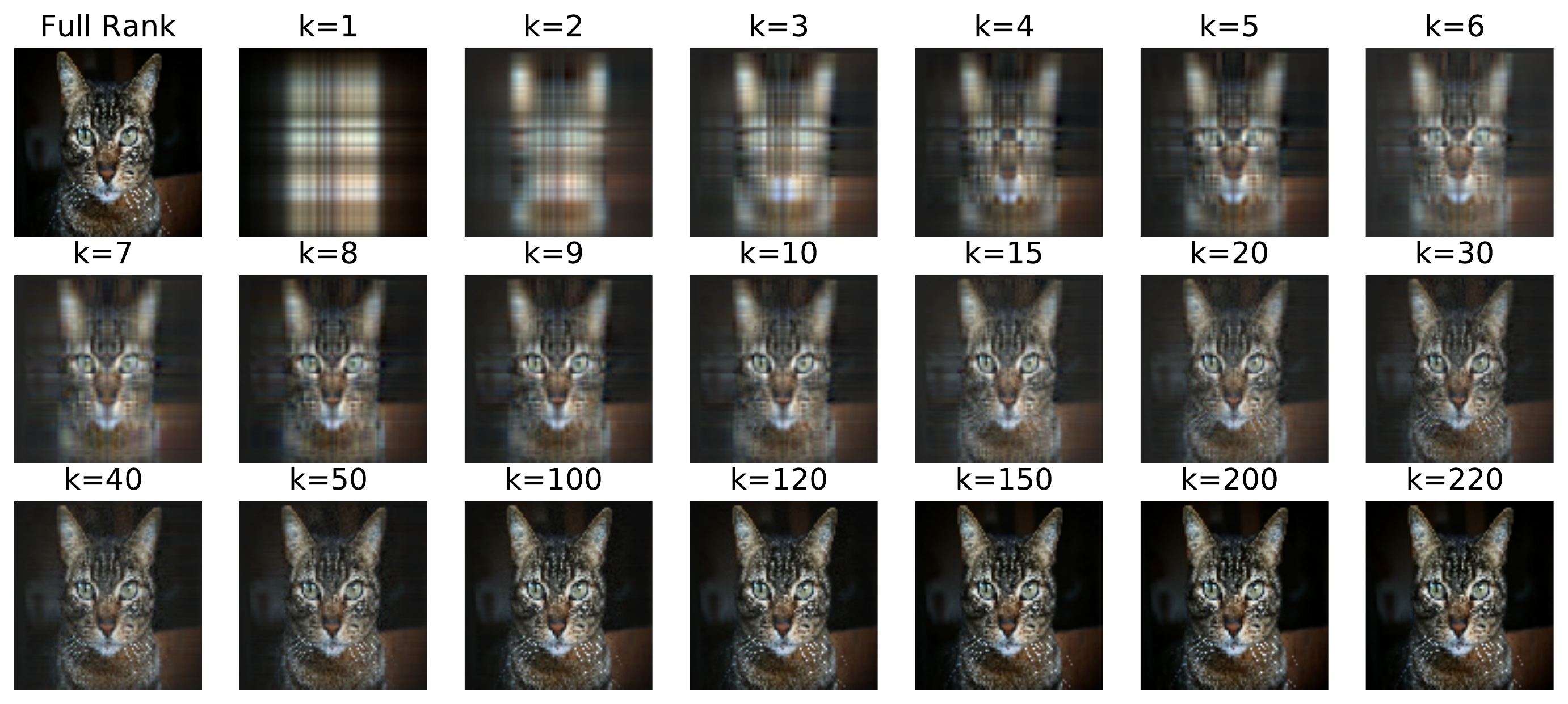}
    \caption{Low-rank approximations of the same image. Transitioning from low-rank approximations of images to higher-rank approximations yields better image quality.}
    \label{fig:low_rank_approximations}
\end{figure*}

\subsection{Notation}

We consider a neural network $f( \cdot)$ used for classification where $f(x)_{i}$ represents the softmax probability that image $x$ corresponds to class $i$. Images are represented as $x \in
[0,1]^{w \times h \times c}$, where $w, h, c$ are the width, height and number of channels of the image. We denote the classification of the network as
$r(x) = \argmax_{i} f(x)_{i}$, 
with $r^{\ast}(x)$ 
representing the ground truth of the image. 
Given an image $x$ and an $L_{p}$ norm bound $\epsilon$, an adversarial sample $x' = x + \delta$ has the following properties:
\vspace{-0.1cm}
\begin{itemize}
    \item For a perturbation $\delta \in [0,1]^{w \times h \times c}$ added to an image $x$ such that $x' = x + \delta$, $L_{p}(\delta) = {(\sum_{i=1}^{h} \sum_{j=1}^{w} \left|\delta_{i,j}\right|^{p})}^{1/p} \leq \epsilon$ where $p=(1,2,\infty)$.\vspace{-0.1cm}
    
    \begin{itemize}
        \item $p=1$ is the Manhattan norm, defined as the sum of the absolute values of $\delta$.\vspace{-0.1cm}
        \item $p=2$ is the Euclidean norm of $\delta$.\vspace{-0.1cm}
        \item $p=\infty$ is the infinity norm or max-norm of $\delta$, defined as the largest absolute value in $\delta$. \vspace{-0.2cm}
    \end{itemize}

    \item $r(x') \neq r^{\ast}(x) = r(x)$. This means that the prediction on the adversarial sample is incorrect while the original prediction is correct.
\end{itemize}

\subsection{Adversarial Samples}

In this work we consider adversaries with white-box access to the neural network. In the white box threat model all information about the neural network is accessible. Using this information, adversaries can compute gradients with respect to inputs by backpropagation. White box attacks can be either targeted or untargeted. In targeted attacks, adversaries seek to generate an adversarial sample $x'$ from an image $x$ to force the neural network $f(x)$ to predict a pre-specified target $t$ that is different from the true class $r^{*}(x)$, while in untargeted attacks adversaries seek to find an adversarial sample $x'$ whose prediction is simply different from that of the true class.

Numerous methods to generate adversarial samples have been proposed~\cite{moosavi2016deepfool, goodfellow2014explaining, carlini2017adversarial, universal, madry2018towards}. In this work, we focus on the PGD attack with random starts, which has been shown to be an effective universal first-order adversary against neural networks~\cite{madry2018towards}. For a neural network $f$, PGD is an iterative adversarial attack method that seeks to generate a targeted adversarial sample $x'$ from an original image $x$ with maximum perturbation limit $\epsilon$. At each iteration, it performs a gradient descent step in the loss function w.r.t the image pixel values and the target class $t$ and projects the perturbed image onto the feasible space, which is either a maximum per-pixel perturbation of $\epsilon$ (for $L_{\infty}$ perturbations) or a maximum Euclidean perturbation distance from $x$ of $\epsilon$ (for $L_{2}$ perturbations).

\noindent \textbf{Adversarial training.} Adversarial training defends against adversarial samples by training networks on adversarial perturbations that are generated on-the-fly. Adversarial training with $L_{\infty}$ PGD samples has been shown to be among the most effective methods in mitigating these attacks~\cite{xie2019feature, madry2018towards}.


\subsection{Explaining Adversarial Samples}
\noindent Recent work has begun to understand the origin of adversarial samples. Ilyas et al. demonstrate that models trained on adversarial samples can generalize to test sets~\cite{ilyas2019adversarial}, and posit that adversarial samples are generalizable features that neural networks learn which are invisible to humans. Yin et al. and Wang et al.~\cite{fourier_perspective, wang2020high} propose that adversarial samples for non-robust neural networks are in the high-frequency domain.
Jere et al.~\cite{jere2019principal} hypothesize that adversarial samples require significantly more principal components of an image to reach the same prediction compared to natural images. Our work is most similar to that of Yin et al.~\cite{fourier_perspective}, in that we observe naturally trained CNNs are sensitive to higher-rank features, and that adversarial training makes them more biased to low-rank features. We explore the relationship between Fourier and low-rank features in the appendix.

\subsection{Feature Attribution Methods}

The problem of feature attribution seeks to \textit{attribute the prediction of deep neural networks to its input features}. Most methods of feature attribution involve variants of visualizing the gradients $\frac{\partial f}{ \partial x}$ of the network with respect to the top predicted class $i$~\cite{selvaraju2017grad, shrikumar2017learning, binder2016layer}. A significant challenge in designing attribution techniques is that attributions without respect to a fixed baseline are hard to evaluate, a problem which was successfully addressed by Integrated Gradients~\cite{sundararajan2017axiomatic}. In Integrated Gradients, the baseline for an image is established with respect to a completely black image $\Tilde{x}$, and a straight line is defined from $\Tilde{x}$ to $x$ with increasing brightness values. Gradients are weighed and computed at each of these steps to result in the final saliency map, which is mathematically equivalent to the path integral along a straightline path from the baseline to the input. In our work, we perform a similar evaluation with our baseline set by the minimum \textit{rank} of an image, and with gradients computed along increasing image rank. Further details can be found in Section~\ref{sec:RIG}, with numerous examples of RIG saliency maps in Figures~\ref{fig:RIG},~\ref{fig:RIG2} and in the appendix.

\subsection{Low-rank approximations}
\noindent Low-rank representations of matrices (obtained via the Singular Value Decomposition) can capture a significant amount of information while simultaneously eliminating spurious correlations, and have recently been used in several Deep Learning applications, such as compression of CNN filters~\cite{denton2014exploiting} and compression of internal representations of attention matrices in transformers~\cite{choromanski2020rethinking, wang2020linformer}.

\section{Methodology} ~\label{sec3}
\vspace{-0.4cm}

In this section we first introduce the theoretical basis behind singular value decomposition, followed by the algorithm used in the rest of the paper to perform low-rank decomposition of RGB images.

\subsection{Eigendecomposition of images}
Eigendecomposition is commonly used to factor matrices into a canonical form, where the matrix is represented in terms of its eigenvalues and eigenvectors. In this work, we focus on utilizing the Singular Value Decomposition (SVD) to obtain low-rank approximations of an input to a neural network. In particular, let an image $x \in [0,1]^{w \times h \times c}$, where $w, h, c$ are the width, height and number of channels of the image respectively, and $w \leq h$. For each channel $m \in {1,2,\ldots c}$ the singular value decomposition on the matrix $A \in [0,1] ^{w \times h}$ yields:

\begin{equation}
    A = U \Sigma V^{T}
\end{equation}

$U$ and $V$ are orthonormal matrices, and $\Sigma$ is a $w \times h$ ($w \leq h$) diagonal matrix with entries $(\sigma_{1}, \sigma_{2}, \ldots , \sigma_{w})$ denoting the singular values of $A$ such that $\sigma_{1} \geq \sigma_{2}, \ldots \geq  \sigma_{w} \geq 0 $. According to the Eckart-Young-Mirsky theorem, the best $k$ rank approximation to the matrix $A$ in the spectral norm $||\cdot||_{2}$ is given by:
\begin{equation}
    A_{k} = \sum_{j=1}^{k} \sigma_{j} u_{j} v_{j}^T
\end{equation}
where $u_{j}, v_{j}$ denote the $j$th column of $U$ and $V$ respectively. This process is also termed as \textit{truncated SVD}. For an $n$-rank matrix $A$, its $k$-rank approximation can also be expressed as $A_k = U \Sigma_{k} V^{T}$, where $\Sigma_{k}$ is constructed from $\Sigma$ by setting the smallest $n-k$ diagonal entries set to zero. 

\subsection{Algorithm}
We define the top class $r(x) = \argmax_{i} f(x)_{i}$ as the predicted class on the full-rank image $x$. For an image $x \in [0,1]^{w \times h \times c}$, we perform singular value decomposition for each color channel $c$, reconstruct a low-rank approximation using $k$ singular values, and perform inference on the rank-$k$ image. Algorithm ~\ref{alg:algorithm_truncation} highlights the steps that occur, and Figure~\ref{fig:low_rank_generation_methodology} illustrates the steps involved in generating a rank-$k$ image by truncating each of the RGB channels and reconstructing the image. Experimental results regarding latency and runtime can be found in the appendix.

\begin{algorithm}[ht!]
\SetAlgoLined
\KwResult{Accuracies for each rank $k=0:w$}
$full\_rank\_preds \leftarrow f(x_{1:N})$ \\
$rank\_k\_acc \leftarrow zeros(w+1)$ \\
\For{$k = 0:w$}{
$rank\_k\_x = zeros\_like(x_{1:N})$ \\
    \For{$i = 1:N$} {
        \For{$channel = 1:c$} {
            $u, \sigma, v = SVD(x[i][channel])$ \\
            $\sigma[k:w] = 0$ \\
            $rank\_k\_x[i][channel] = u \ diag (\sigma) \ v$ \\
        }
    }
    $rank\_k\_acc[k] = (f(rank\_k\_x) == full\_rank\_preds)$
}
\Return $rank\_k\_acc$
\caption{Finding the accuracies of a model $f(\cdot)$ for a batch of images $x_{1:N}$, where $x_i \in [0,1]^{w \times h \times c}$ as a function of the input rank.}
\label{alg:algorithm_truncation}
\end{algorithm}

\vspace{-0.1cm}
\begin{figure}[ht!]
    \centering
    \includegraphics[width=0.4\textwidth]{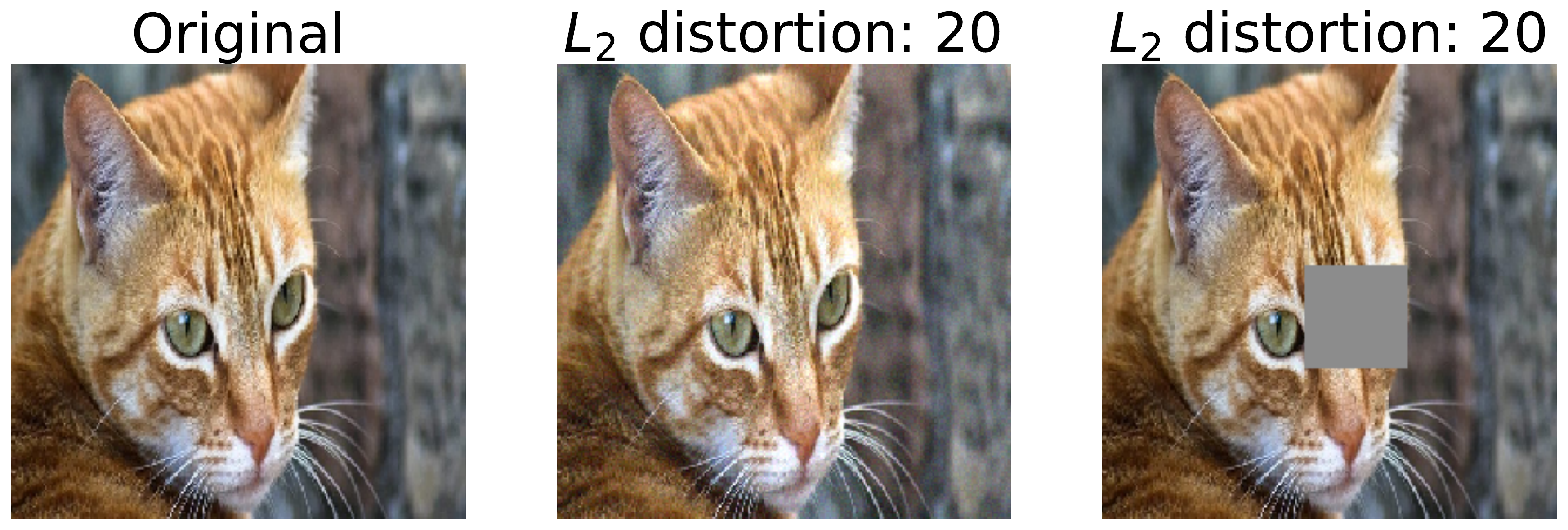}
    \caption{Limitations of $L_{p}$ distortions. The middle image is generated by adding uniform noise to the first one. Both distorted images have identical $L_{2}$ distortions, but are perceptually dissimilar. The patched image (far right) represents one of a near infinite number of possible images with identical $L_2$ distortions as the middle image.}
    \label{fig:distorted_kitty}
\end{figure}
\vspace{-0.3cm}

\begin{table}[ht!]
\centering
\begin{tabular}{c|c|c}
\hline
\textbf{Dataset} & \textbf{\begin{tabular}[c]{@{}c@{}}Naturally\\ trained\\ Models\end{tabular}} & \textbf{\begin{tabular}[c]{@{}c@{}}Robust \\ models\end{tabular}} \\ \hline
ImageNet & \begin{tabular}[c]{@{}c@{}}ResNet-50,\\ VGG-19,\\ DenseNet-201\end{tabular} & \begin{tabular}[c]{@{}c@{}}ResNet-50 \\ ($L_{\infty}=4/255, 8/255$),\\ ($L_{2}=3.0$)\end{tabular} \\ \hline
CIFAR-10 & ResNet-50 & \begin{tabular}[c]{@{}c@{}}ResNet-50\\ ($L_{\infty}=8/255$),\\ ($L_{2}=0.25,0.5,1.0$)\end{tabular} \\ \hline
\end{tabular}
\caption{Experimental setup to investigate behavior of benign v/s adversarially robust neural networks.}
\vspace{-0.4cm}
\label{tab:experimental_setup}
\end{table}

\section{Towards robustness metrics beyond $L_{p}$ distortions} ~\label{sec4}
In this section we highlight several limitations of $L_{p}$ distortions, followed by experimental results for naturally trained and adversarially robust CNNs. Finally, we introduce Rank-Integrated Gradients.

\subsection{Limitations of $L_{p}$ distortions}

Extensive experiments have been conducted to secure neural networks against $L_{p}$-norm bounded perturbations, such as adversarial training~\cite{madry2018towards,kannan2018adversarial,xie2019feature,shafahi2019adversarial} and certified defenses against adversarial samples~\cite{raghunathan2018certified,wong2017provable,zhang2019theoretically,cohen2019certified}. Unfortunately, $L_{p}$-distortions represent a small fraction of potential image modifications. An infinite number of modified images exist that possess identical norm-bounded perturbations with respect to a base image. Furthermore, identical $L_p$ norm-bounded distortions may be extremely different perceptually, pointing to $L_{p}$-norm robustness potentially being misaligned with human perception (Figure~\ref{fig:distorted_kitty}). Based on these observations and limitations, we argue that image rank and rank-based robustness metrics might be better suited to capture image modifications than $L_{p}$-distortions for the following reasons.

\begin{itemize}

    \item Matrix rank for an image $x \in [0,1]^{w \times h \times c}$ ($w \leq h$) is restricted to the set of integers ${1,2,...,w}$. The set of images generated by rank truncation is a bijective mapping from rank $k$ to generated images $x_{k}$. This is in contrast to images generated with $L_{p}$ distortions whose mapping contains infinite possibilities.
    
    \item Low-rank image approximation effectively captures a much larger range of image modifications (Figure~\ref{fig:low_rank_approximations}) that is more perceptually aligned with human vision that might not be captured with $L_{p}$ distortions. Transitioning from low-rank approximations of images (unrecognizable to humans) to higher-rank approximations (recognizable to both DNNs and humans) effectively allows us to better understand the gap between human and computer vision.
    
\end{itemize}

\subsection{Rank Dependence of CNNs}

\begin{figure}
    \centering
    \includegraphics[width=0.4\textwidth]{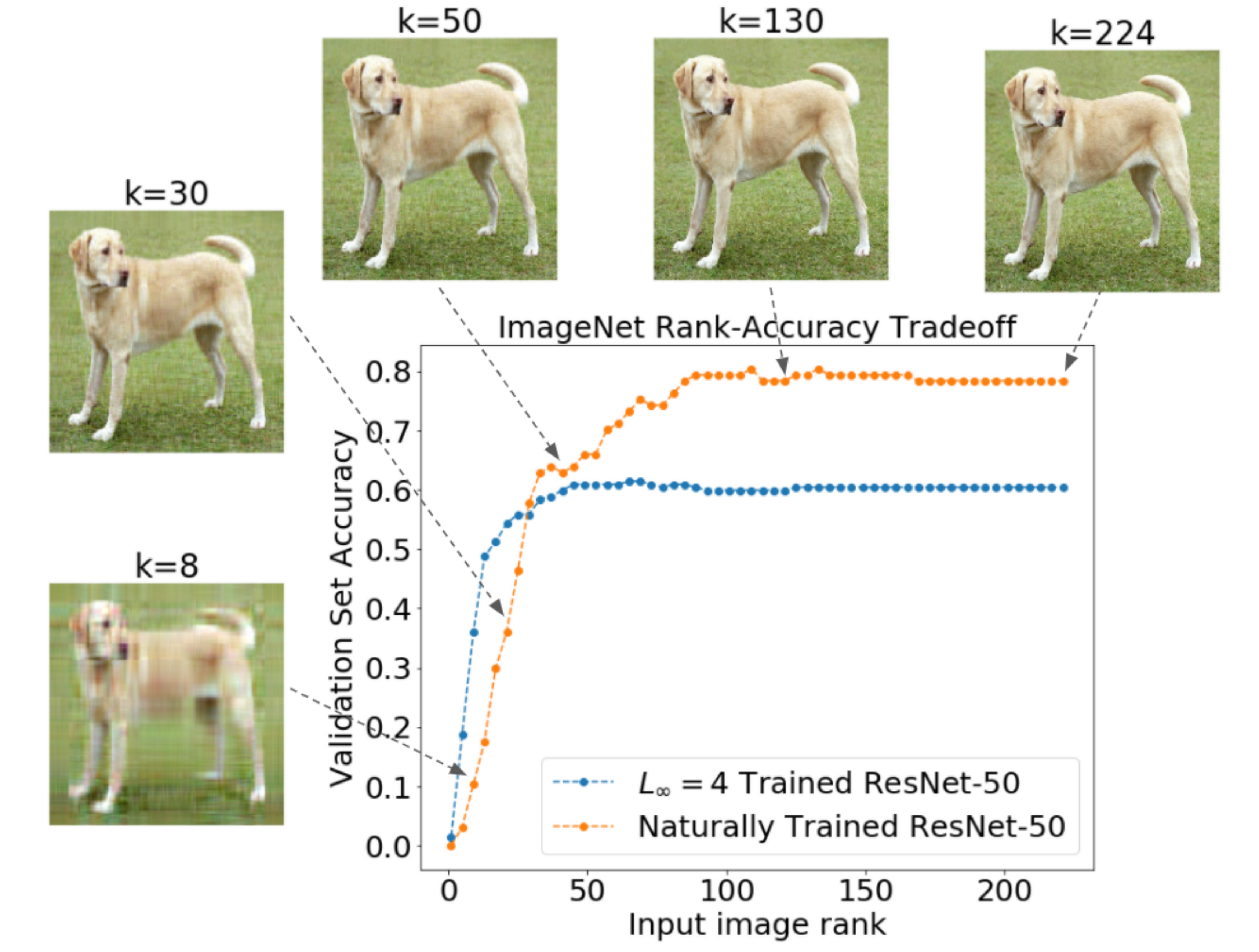}
    \caption{Dependence of test accuracy of naturally trained and $L_\infty=4/255$ robust ResNet-50 models on input rank of natural images (subset of ImageNet validation images). }
    \vspace{-0.4cm}
    \label{fig:imagenet_resnet_50_rank_distorted}
\end{figure}


We seek to understand the dependence of image rank and classifier accuracy for ResNet-50. We observe that naturally trained and robust CNNs use features that are highly different in their rank properties. Particularly, naturally trained CNNs use high-rank features that are often invisible to humans, while robust CNNs do not respond to these features but rather rely on highly visible low-rank features. For both models we randomly sampled $1000$ images from the ImageNet validation set and cropped and reshaped each image to have a shape of $(224 \times 224 \times 3)$. For each image, we performed low-rank approximation for every possible rank prior to inference according to Algorithm~\ref{alg:algorithm_truncation}. \vspace{-0.2cm}

\subsubsection{Behavior of Naturally Trained CNNs}

\begin{figure*}[ht!]
    \centering
    \begin{subfigure}[b]{0.33\textwidth}
        \centering
           \includegraphics[width=\textwidth]{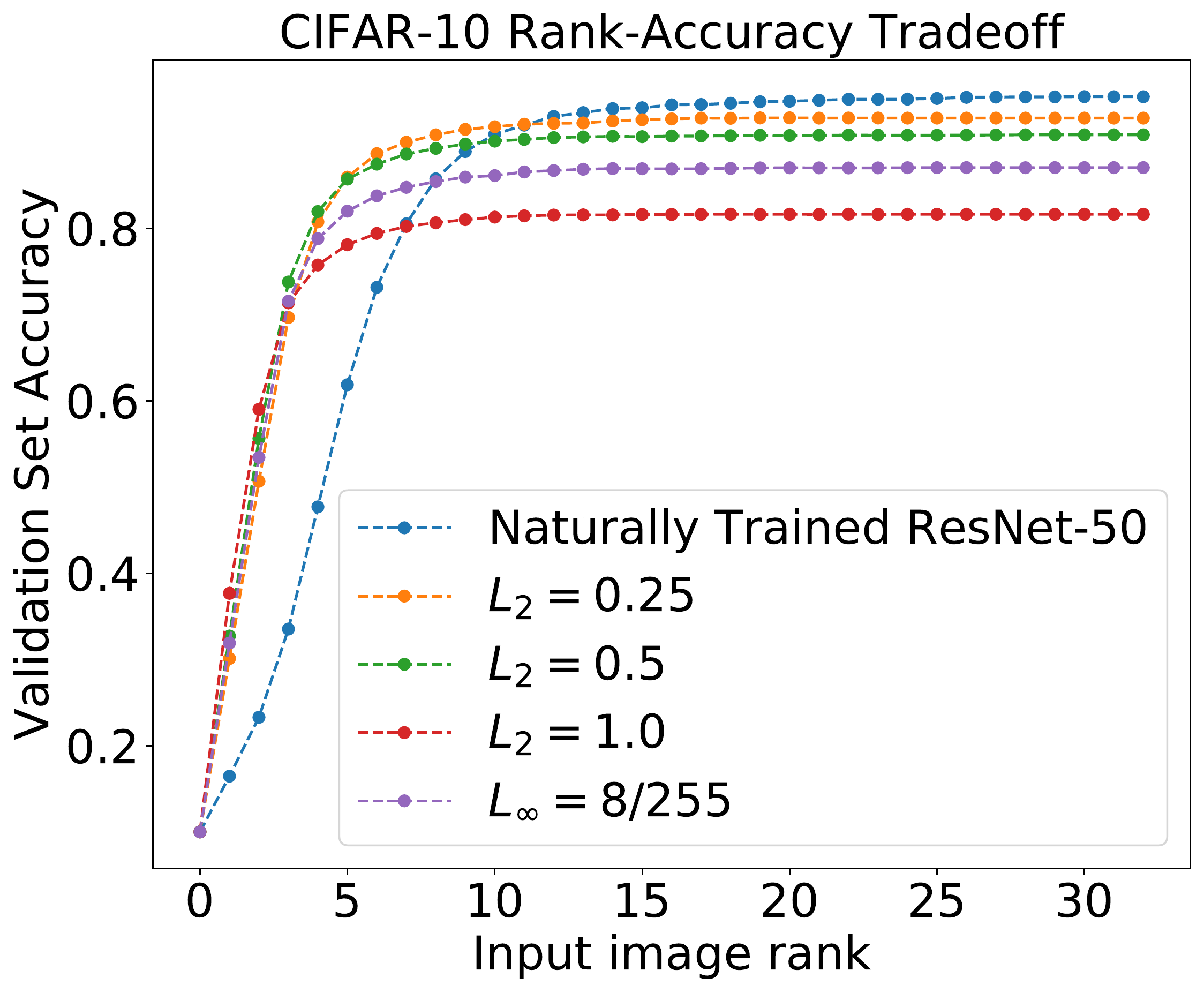}
        \caption{}
        \label{fig:cifar_10_rank_spectrum}
    \end{subfigure}
    \hspace{20pt}
    \begin{subfigure}[b]{0.33\textwidth}
        \centering
        \includegraphics[width=\textwidth]{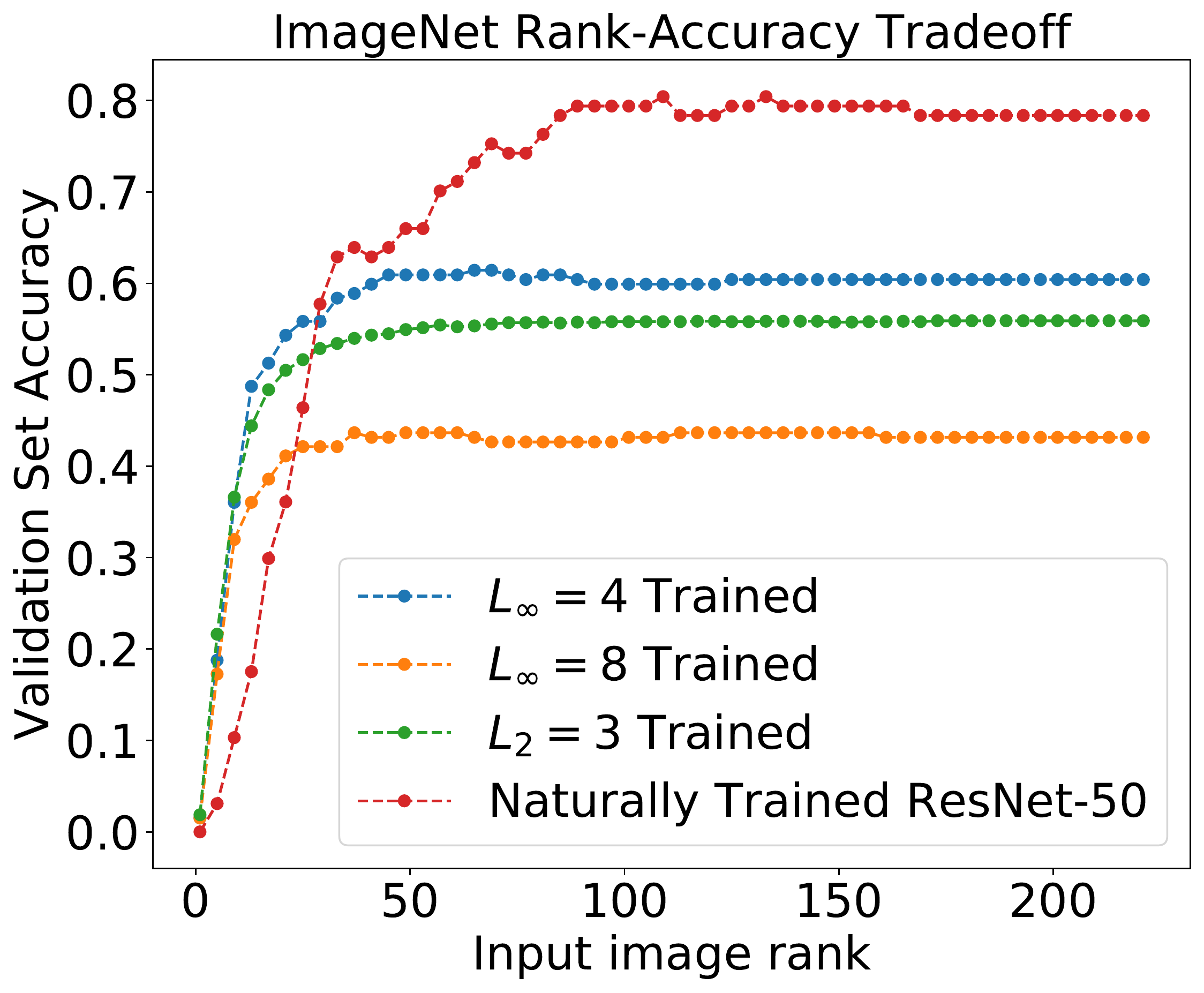}
        \caption{}
        \label{fig:imagenet_rank_spectrum}
    \end{subfigure}
    \hspace{20pt}
    \begin{subfigure}[b]{0.33\textwidth}
        \centering
           \includegraphics[width=\textwidth]{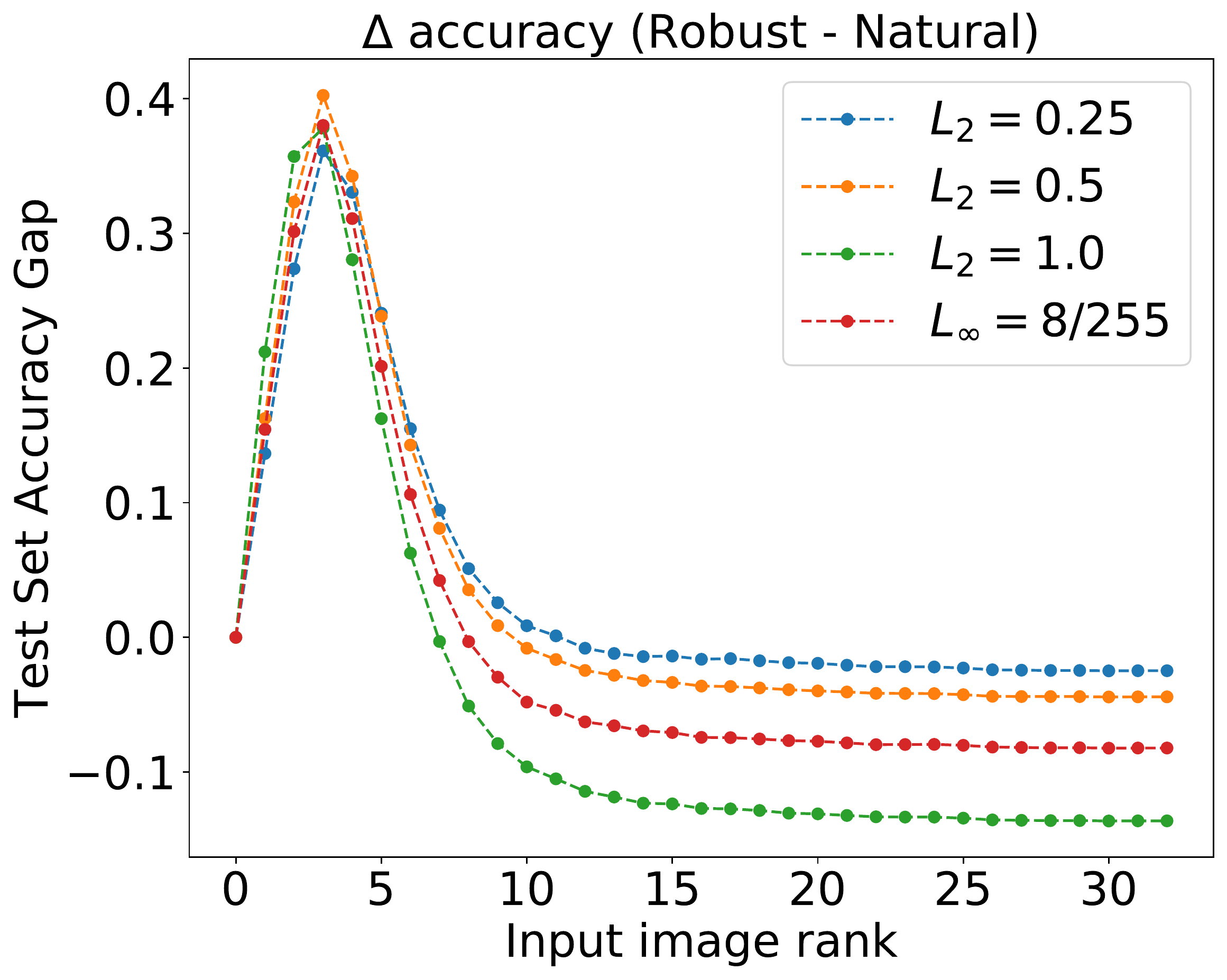}
        \caption{}
        \label{fig:cifar_10_rank_spectrum_gap}
    \end{subfigure}
    \hspace{20pt}
    \begin{subfigure}[b]{0.33\textwidth}
        \centering
        \includegraphics[width=\textwidth]{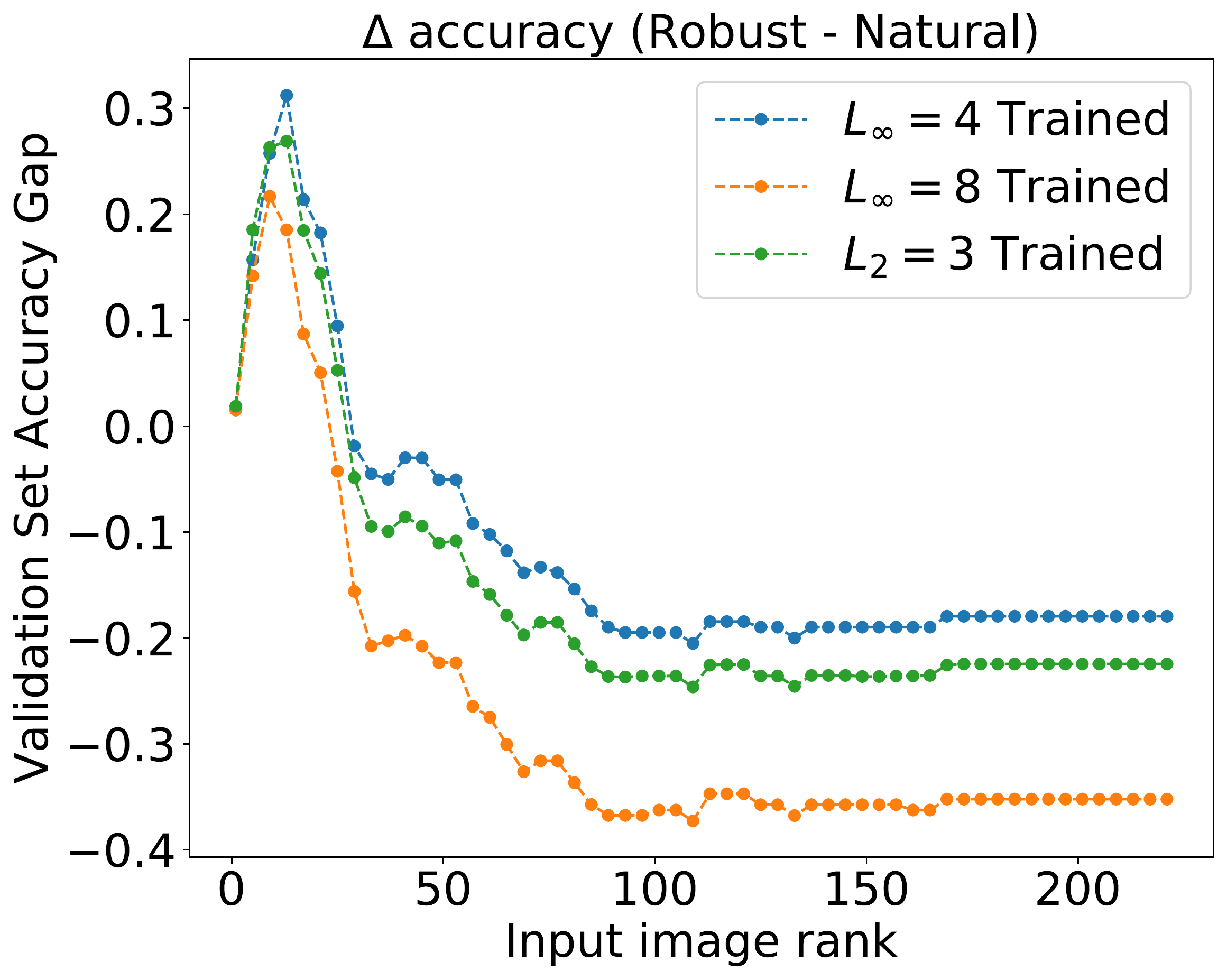}
        \caption{}
        \label{fig:imagenet_rank_spectrum_gap}
    \end{subfigure}
    \caption{(a) CIFAR-10 rank spectrum for naturally trained and robust ResNet-50 models (b)  ImageNet rank spectrum for naturally trained and adversarially robust ResNet-50 models. (c) Gap in test accuracy between robust and naturally trained CIFAR-10 ResNet-50 models. (d) Gap in test accuracy between robust and naturally trained ImageNet ResNet-50 models.}
\end{figure*}

We investigate the behavior of the ImageNet-trained ResNet-50~\cite{he2016deep}, VGG-19~\cite{simonyan2014very} and DenseNet-201~\cite{huang2017densely} (Table~\ref{tab:experimental_setup}) CNN architectures trained on the ImageNet dataset~\cite{imagenet_classification_first}
Experimental results for the VGG-19 and DenseNet models can be found in the Appendix. In Figure~\ref{fig:imagenet_resnet_50_rank_distorted} we observe the top-1 accuracy for ResNet-50 (orange) on these truncated images for both naturally trained and robust models. We make the following notable observations for this as well as for Figures~\ref{fig:cifar_10_rank_spectrum} and ~\ref{fig:imagenet_rank_spectrum}: 
\begin{itemize}
    \item Classifier accuracy sharply increases for lower-ranked images (rank-$50$ to rank-$100$) followed by saturation around rank-$100$ for ImageNet trained models. We observe similar behavior at rank-$15$ for CIFAR-10 trained models (Figure~\ref{fig:cifar_10_rank_spectrum}).
    \item We observe that the features corresponding to this increase in accuracy, namely rank-$50$ to rank-$100$, contribute no meaningful semantic content to the image (Figure~\ref{fig:low_rank_approximations}), indicating that naturally trained CNNs exploit features that are often invisible to humans. (~\ref{fig:imagenet_rank_spectrum})
    
\end{itemize}
\vspace{-0.4cm}





\subsubsection{Behavior of Adversarially Robust CNNs}

We investigate the behavior of pretrained adversarially robust CNNs from the robustness library~\cite{robustness} as highlighted in column 2 of Table~\ref{tab:experimental_setup}. 
Our results for an $L_{\infty}=4/255$ robust ResNet-50 model can be seen in Figure~\ref{fig:imagenet_resnet_50_rank_distorted} (blue). In contrast to naturally trained models we observe very different behaviors for robust models. Notably, we observe that:
\vspace{-0.2cm}
\begin{itemize}
    \item Robust CNN accuracy for full-rank images is lower than that of naturally trained CNNs for both ImageNet and CIFAR-10 trained models (which has been observed in previous work~\cite{tsipras2018robustness}). 
    
    \item Robust CNN accuracy for lower-rank images is higher than that of naturally trained CNNs. Surprisingly, it is significantly superior for lower-ranked images, with a $>20\%$ validation set accuracy improvement in both datasets (Figure~\ref{fig:cifar_10_rank_spectrum_gap} and~\ref{fig:imagenet_rank_spectrum_gap}).
    
    \item Robust CNN accuracy increases much more quickly than naturally trained CNNs for lower-rank images, and does not exhibit the same dependence on features between rank-50 and rank-100 for the ImageNet dataset.
\end{itemize}

The rank-accuracy tradeoff as well as superior performance of robust models for lower-rank images has not been observed before, and to the best of our knowledge ours is the first work to identify such phenomena.
We further observe in Figures~\ref{fig:cifar_10_rank_spectrum} and~\ref{fig:imagenet_rank_spectrum} that this rank behavior persists across CNNs trained with different $L_{\infty}$ bounds, different $L_{p}$-norm metrics and different datasets. 
While there exist minor differences between the rank behavior of $L_{\infty}$ and $L_{2}$ robust CNNs, their behaviors are largely distinct from those of naturally trained CNNs.

\begin{table*}[ht!]
\centering
\begin{tabular}{c|c|c|c|c|c|c|c}
\hline
 & \multicolumn{3}{c|}{\textbf{CIFAR-10}} & \multicolumn{4}{c}{\textbf{ImageNet}} \\ \hline
\textbf{\begin{tabular}[c]{@{}c@{}}Trained\\  on\end{tabular}} & \textbf{\begin{tabular}[c]{@{}c@{}}Attack \\ success rate\end{tabular}} & \textbf{\begin{tabular}[c]{@{}c@{}}Recovery\\  rate\end{tabular}} & \textbf{\begin{tabular}[c]{@{}c@{}}Top-1\\ accuracy\end{tabular}} & \textbf{\begin{tabular}[c]{@{}c@{}}Trained\\  on\end{tabular}} & \textbf{\begin{tabular}[c]{@{}c@{}}Attack\\  success rate\end{tabular}} & \textbf{\begin{tabular}[c]{@{}c@{}}Recovery\\  rate\end{tabular}} & \textbf{\begin{tabular}[c]{@{}c@{}}Top-1\\ accuracy\end{tabular}} \\ \hline
\textbf{\begin{tabular}[c]{@{}c@{}}Full rank\end{tabular}} & 99.93\% & 0.03\% & 95.21\% & \textbf{Full rank} & 95.87\% & 0.01\% & 78.35\% \\ \hline
\textbf{20} & 99.70\% & 0.15\% & 95.41\% & \textbf{100} & 81.81\% & 7.07\% & 73.99\% \\ \hline
\textbf{10} & 99.43\% & 0.19\% & 94.90\% & \textbf{50} & 73.99\% & 8.57\% & 70.16\% \\ \hline
\textbf{5} & 99.27\% & 0.34\% & 91.54\% & \textbf{30} & 73.19\% & 5.15\% & 69.07\% \\ \hline
\end{tabular}
\caption{Robustness of low-rank CIFAR-10 and ImageNet-trained ResNet-50 models. Attack success rate and recovery rate measured for targeted $20-$step PGD attacks with $L_{\infty}-$bounds of $4/255$. Top-1 accuracy is measured for full-rank test sets.}
\label{tab:robustness_measurement_joint}
\end{table*}

\subsection{Rank Integrated Gradients (RIG)} ~\label{sec:RIG}
\vspace{-0.02cm}
Based on these observations, we seek to visualize the rank-dependency of CNNs. Generating visual explanations for CNN image classifiers typically involves computing saliency maps that take the gradient of the output corresponding to the correct class with respect to a given input vector such as GradCAM and guided-backprop~\cite{simonyan2013deep, selvaraju2017grad, springenberg2014striving}. However, such methods often only capture local explanations for a given image, and are not robust to perturbations to the original image. Other methods involve training simpler, more interpretable surrogate models~\cite{lundberg2017unified, ribeiro2016should} to understand model predictions in a local neighborhood around a given input, but these cannot sufficiently capture rank-based image modifications nor scale to models such as ResNets. 

\begin{figure}[ht!]
    \centering
    \includegraphics[width=0.45\textwidth]{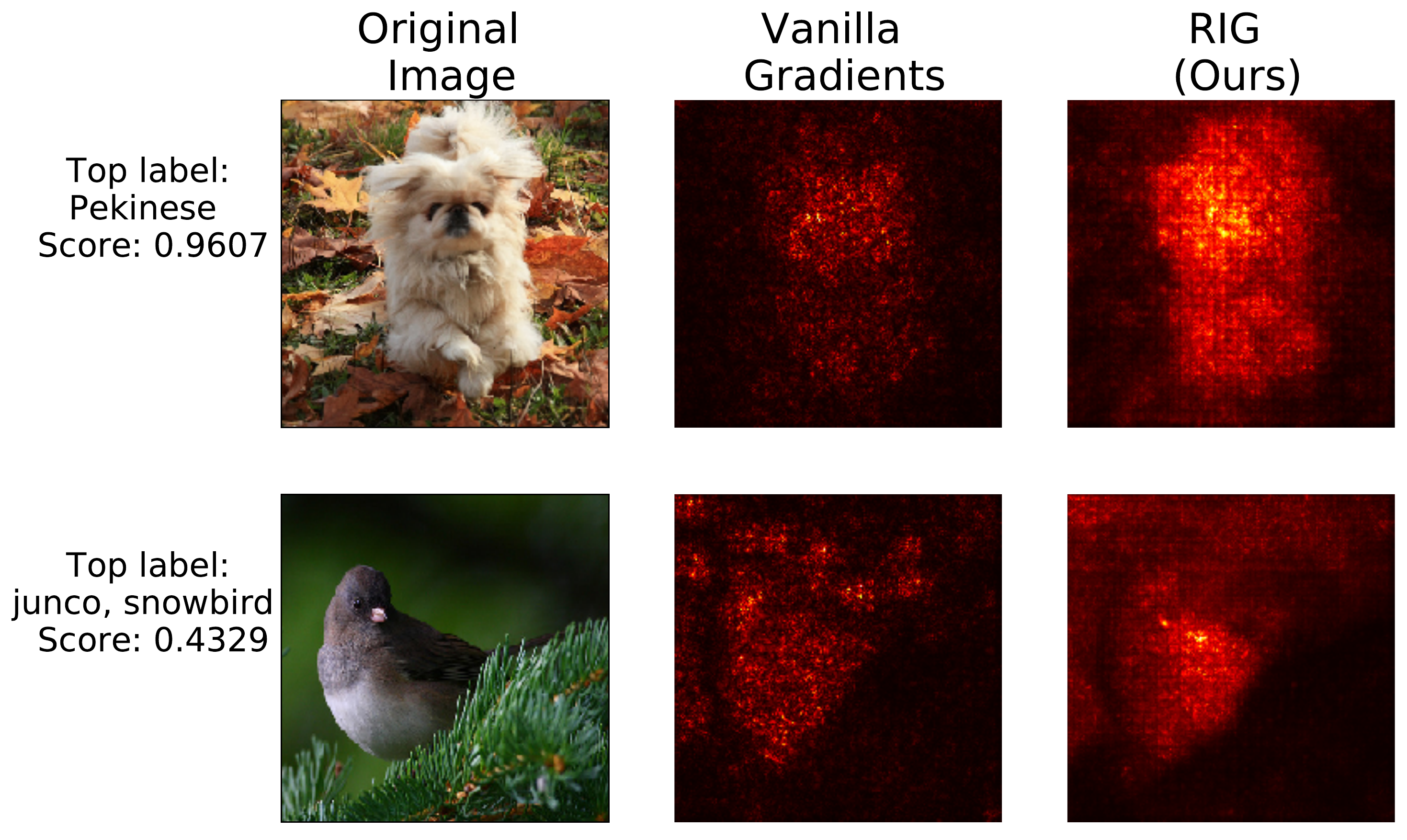}
    \caption{Rank Integrated Gradient Images. In the second row, vanilla gradients highlight features in the background that contribute to the top class but have no meaningful semantic content. RIG highlights these features as well as object specific features, such as the bird's beak.}
    \label{fig:RIG2}
\end{figure}

Feature Attribution methods that seek to capture network predictions while being invariant to perturbations and implementations of the method~\cite{sundararajan2017axiomatic} have been successful in capturing such attributes among image pixels. In particular, Blur Integrated Gradients (BIG)~\cite{xu2020attribution} has been effective in capturing feature attributes through Gaussian-blurred versions of the original image. In this regard our work is most similar to BIG, but we differentiate ourselves in calculating gradients through low-rank representations of an image rather than gaussian-blurred representations.

Intuitively, our technique weighs low-rank representations with their contributions to the gradients of an input for the top class predicted on the full-rank input. Formally, let us denote a classifier $f$ and an input signal $x \in [0,1]^{w \times h \times c}$ where $w \leq h$, with $x_{k}$ as a rank-$k$ image obtained by Algorithm~\ref{alg:algorithm_truncation}. Let us denote an image classifier $f$, and the top predicted class $i$ for a full rank image $x_{w}$. Let $(\frac{\partial f(x_k)}{\partial x_k})_{i}$ denote the maximum gradient across all color channels $c$. Then, our method computes RIG as:
\vspace{-0.08cm}
\begin{equation}
    RIG(x,f,i) =  \sum_{k=1}^{w} \frac{w - k}{w} \times  (\frac{\partial f(x_k)}{\partial x_k})_{i}
\end{equation}

\noindent RIG requires no modification to the model and is extremely easy to implement, requiring less than $10$ lines of PyTorch code and using a few calls to the gradient operation, thereby allowing even novice practitioners to easily apply the technique. Examples can be found in Figure~\ref{fig:RIG} and ~\ref{fig:RIG2}.

\section{Transferability of Rank-Based Features} ~\label{sec5}
\vspace{-0.3cm}

Motivated by the disparities in the behavior between naturally trained and adversarially robust CNNs in Section~\ref{sec4}, we proceeded to test the following hypotheses:
\begin{itemize}
    \item Do CNNs trained solely on high-rank representations generalize to a full-rank test set? 
    \item Do CNNs trained solely on low-rank representations generalize to a full-rank test set? 
    \item Do CNNs trained solely on low-rank representations improve robustness to $L_{p}$-norm bounded attacks?
\end{itemize}


\subsection{Training on solely high-rank representations} \label{subsection:train_high_rank}
\begin{figure}[ht!]
    \centering
    \includegraphics[width=0.45\textwidth]{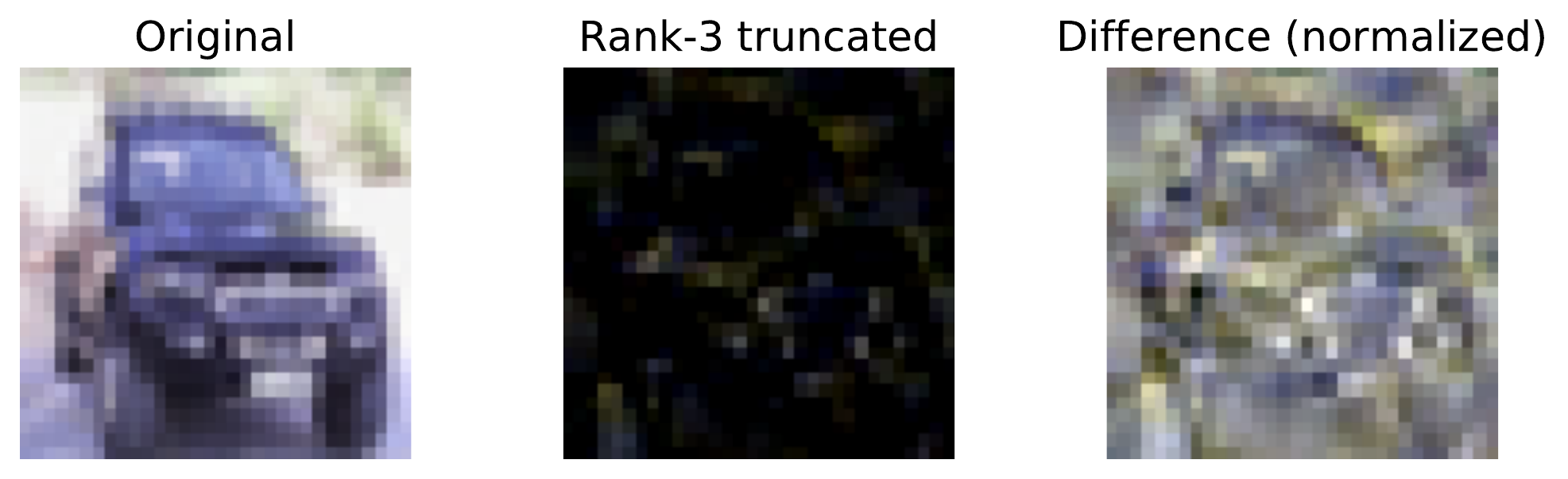}
    \caption{Reverse rank truncated CIFAR-10 image. Middle column corresponds to images where the $3$ largest singular values were set to 0, with the difference between the original image and truncated image on the right column.}
    \vspace{-0.3cm}
    \label{fig:reverse_truncated_rank}
\end{figure}

\vspace{-0.1cm}
We conducted experiments to test the hypothesis: \textit{Do CNNs trained on solely high-rank representations generalize to the test set?} We modified Algorithm~\ref{alg:algorithm_truncation} to zero out the $k$-largest singular values (instead of the $k$-smallest singular values), thereby creating images that consist solely of higher-rank features that are largely imperceptible and difficult to interpret even when visualized (Figure~\ref{fig:reverse_truncated_rank}). We trained the ResNet-50 architecture on a modified version of the CIFAR-10 dataset consisting solely of these higher-ranked representations, and evaluated it on the full-rank CIFAR-10 test sets. Each network was trained for $350$ epochs with an SGD optimizer, with learning rate of $0.1$, momentum $0.9$ and weight decay of $0.0005$. We decreased the learning rate by $10$ after the $150$th and $250$th epochs. 

We observe that for training images where the first $3$ singular values are truncated (Figure~\ref{fig:reverse_truncated_rank}), we get a full-rank test accuracy of $28.02\%$. Such non-random accuracy shows that high-rank representations of an image contain meaningful features for generalization. However, this accuracy quickly decreases as we increase the number of truncated singular values. Further details can be found in the appendix.



\subsection{Training on low-rank representations}
\label{sec:training_low_rank}
We conducted experiments to address the hypothesis: \textit{Do CNNs trained solely on low-rank representations generalize to the test set?} We trained ResNet-50 models for CIFAR-10 and ImageNet solely on low-rank representations and observed their accuracy on the held out full-rank test sets. 
\vspace{-0.3cm}
\subsubsection{CIFAR-10}
\vspace{-0.1cm}
We trained ResNet-50 models on low-rank representations with similar hyperparameters as Section~\ref{subsection:train_high_rank}.
We observe that low-rank representations are sufficient to achieve test accuracy of more than $90\%$ for full-rank CIFAR-10 test sets. Surprisingly, training CIFAR-10 on rank-5 images yields test accuracy of $91.54\%$, which is only $3.67\%$ lower than when trained on a full-rank dataset (Table~\ref{tab:robustness_measurement_joint}). Increasing the rank of training-set images quickly closes this gap, and rank-20 images have almost identical test accuracy to full-rank images, indicating that the high-rank information in images is largely irrelevant to make predictions for CIFAR-10. This corroborates the results we obtained from Figure~\ref{fig:cifar_10_rank_spectrum}, where we observed that test accuracy does not rely on features past rank-15 for CIFAR-10. Further experimental details can be found in the Appendix.
\vspace{-0.3cm}
\subsubsection{ImageNet}
\vspace{-0.1cm}

We trained ResNet-50 using the same hyperparameters as described in the original ResNet-50 paper~\cite{he2016deep} on low-rank versions of the ImageNet dataset. Specifically, we trained on rank-$30,50$ and $100$ representations (Table~\ref{tab:robustness_measurement_joint}). Despite rank-100 and full-rank images being visually identical, we observe that the full-rank validation set accuracy for rank-100 trained ResNet-50 is $4.3\%$ lower than that of ResNet-50 trained on full-rank images. This indicates that the ImageNet data consists of a large number of imperceptible, high-rank features that do not contain semantically meaningful content but contribute to test accuracy.


\subsection{Robustness as an emergent property of low-rank representations}
In this section we conducted experiments to test the hypothesis: \textit{Do CNNs trained solely on low-rank representations improve adversarial robustness to $L_{p}$-norm bounded attacks?} 
To tackle this, we performed $20$-step, $L_{\infty} = 4/255$ PGD adversarial attacks on the low-rank CIFAR-10 and ImageNet-trained ResNet-50 models from Section~\ref{sec:training_low_rank}. Our experimental results can be found in Table~\ref{tab:robustness_measurement_joint}. Notably, we observe that adversarial robustness to $L_{\infty}$ attacks improves with training on lower-ranked image representations for ImageNet. However, this does not hold true for CIFAR-10 trained models. 
Furthermore, strategies such as adversarial training~\cite{madry2017towards} or feature denoising~\cite{xie2019feature} offer superior performance than training on low-rank representations.

\section{Discussion} \label{sec6}

Prior work on interpreting adversarial samples~\cite{ilyas2019adversarial, Wang_2020_Frequency} hypothesized that images consist of \textit{robust} and \textit{non-robust} features, where robust features are largely visible to humans while non-robust features are not. Further work argues that robustness leads to improved feature representations~\cite{salman2020adversarially}. Our findings appear to support these claims. Specifically, we observe that due to their large contribution to image quality and predictive performance for robust networks, low-rank features are analogous to \textit{robust} features and can be generated through low-rank truncation. Conversely, higher-ranked features which do not contribute to robust network predictions are analogous to \textit{non-robust} features. Furthermore, we observe that quantifying network robustness through $L_{p}$ perturbations does little to capture the massive range of possible image modifications, and often runs into the issue of multiple  perceptually different images having identical $L_{p}$ distortions. Rank-based image modifications simultaneously capture a much larger range of image modifications while offering a $1-1$ mapping from modification parameter to perceptual representation.

With respect to feature attribution, we observe that saliency maps that leverage rank information in images are much more aligned with human vision than conventional vanilla gradients, and offer a new lens into understanding the inner workings of these image classifiers. We hypothesize that there exist several other similar forms of matrix decomposition that allow for visualizations that are more perceptually meaningful as well.

\section{Conclusion} \label{sec7}

Closing the gap between computer and human vision is a challenging and an open problem. 
Human vision remains robust under a variety of image transformations, while neural network based computer vision is still fragile to small $L_{p}$-norm limited perturbations, which futhermore do not capture the full range of image modifications. 
We demonstrate the need for robustness metrics beyond these perturbations, and make several arguments in favor of using image-rank (as obtained by SVD) as a potential alternative.
We demonstrate several behavioral differences between naturally trained and adversarially robust CNN classifiers in terms of their generalization that could not be captured in an $L_{p}$-bound framework. Finally, we propose a simple rank-based feature attribution technique that produces gradient visualizations that are much more perceptually informative than saliency maps. 

\section{Acknowledgements} ~\label{sec8}
\noindent This work was supported by the Semiconductor Research Corporation (AUTO TASK 2899.001) and a Defense Advanced Research Projects Agency (DARPA) Techniques for Machine Vision Disruption (TMVD) grant.
{\small
\bibliographystyle{ieee_fullname}
\bibliography{cvpr}
}

\newpage 

\section{Appendix} ~\label{sec9}

\subsection{Relationship between Fourier Features and Rank-based features}

\noindent \textbf{Statement:} We hypothesize that the rank of a matrix obtained from a $(k\times k)$ low-pass filter in the frequency domain is upper bounded by $k$.

\noindent \textbf{Proof.} Let a matrix $X \in [0,1]^{w\times h} = U \Sigma V^{T}$ with rank $r(X) = min(w,h)$ exist in the spatial domain, and let the Fourier transform operation be represented as $F(\cdot)$. Due to its linearity, the Fourier transform of $X$ can be expressed as $F(X) = W X$. Let us denote the low-pass filtering operation with a window of size $k$ as $L$. By definition, the window will have a max rank of $k$. Then, we can express the $k-$window low-pass filtered version of $X$ as $\tilde Y = L W X$ in the frequency domain, and its spatial domain representation as $\tilde X = W^{-1} L W X$. 

The rank of $\tilde X$ can be expressed as $r(\tilde X) = r(W^{-1} L W X)$. Due to the rank property of matrix multiplication, $r(W^{-1} L W X) \leq min(r(W^{-1}), r(L W X))$. Therefore,

\begin{equation}\label{eq:pareto mle2}
\begin{aligned}
    r(W^{-1} L W X) \leq min(r(W^{-1}), r(L W X)) \\ 
    \implies r(L W X) \leq min(r(X), r(L W)) \\
    \implies r(L W) \leq min(r(L), r(W)) = r(L) = k \\
    \implies r(W^{-1} L W X) \leq r(LWX) \leq r(LW) \leq r(L) \leq k
 \end{aligned}
\end{equation}

Thus, $r(\tilde X) \leq  k$.

\subsection{Rank Integrated Gradients}
\begin{figure}[ht!]
    \centering
    \includegraphics[width=0.47\textwidth]{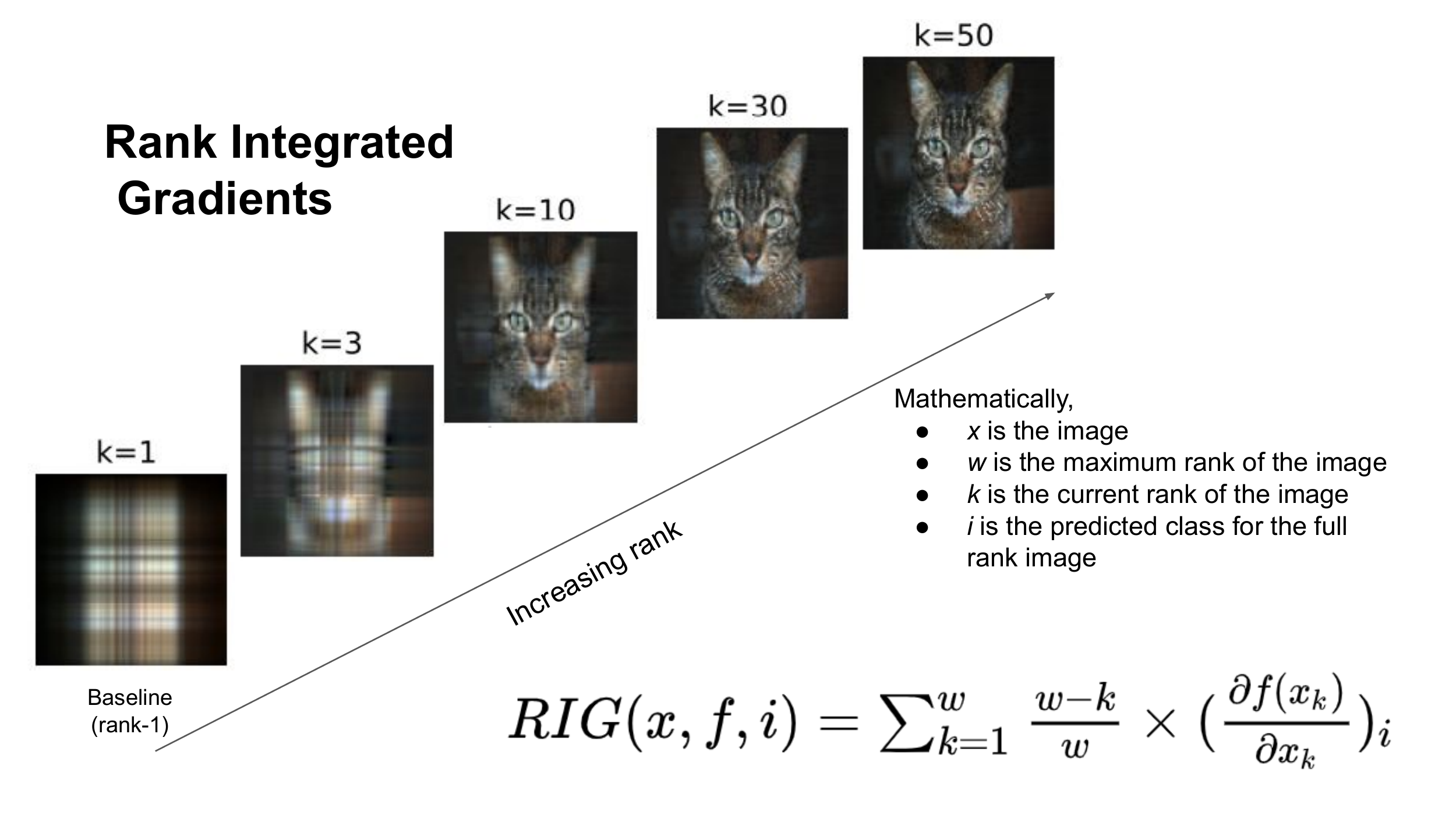}
    \caption{RIG generation.}
    \label{fig:RIG_EXPLAINED}
\end{figure}

\begin{figure}[ht!]
    \centering
    \includegraphics[width=0.47\textwidth]{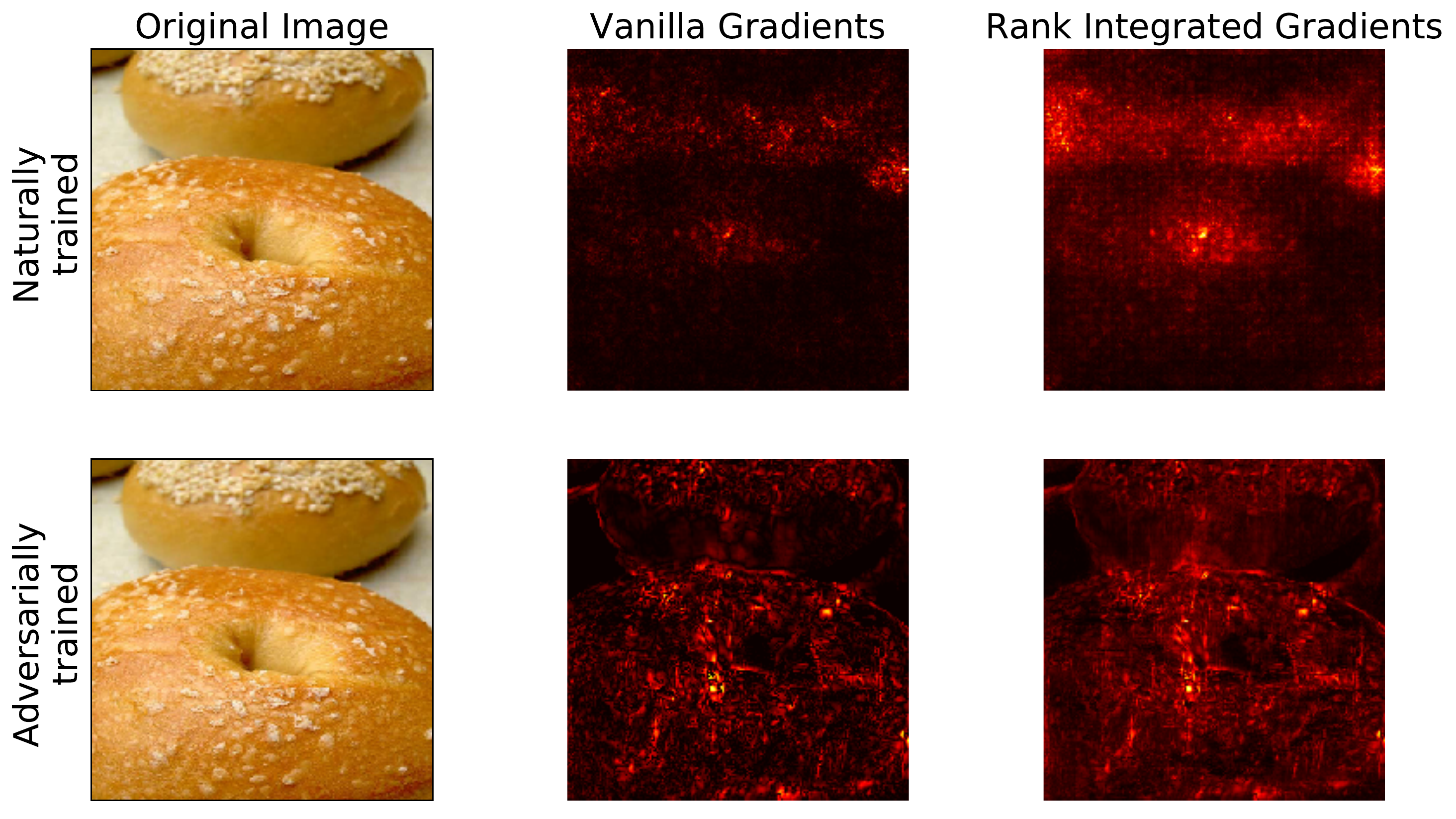}
    \caption{Comparison of RIG for naturally trained and adversarially robust neural networks. Adversarially robust neural networks have image representations that are much more aligned with human perception, which has been previously observed in~\cite{tsipras2018robustness}.}
    \label{fig:RIG_joint}
\end{figure}

\begin{figure*}[ht!]
    \centering
    
    \hspace{20pt}
    \begin{subfigure}[b]{0.7\textwidth}
        \centering
        \includegraphics[width=\textwidth]{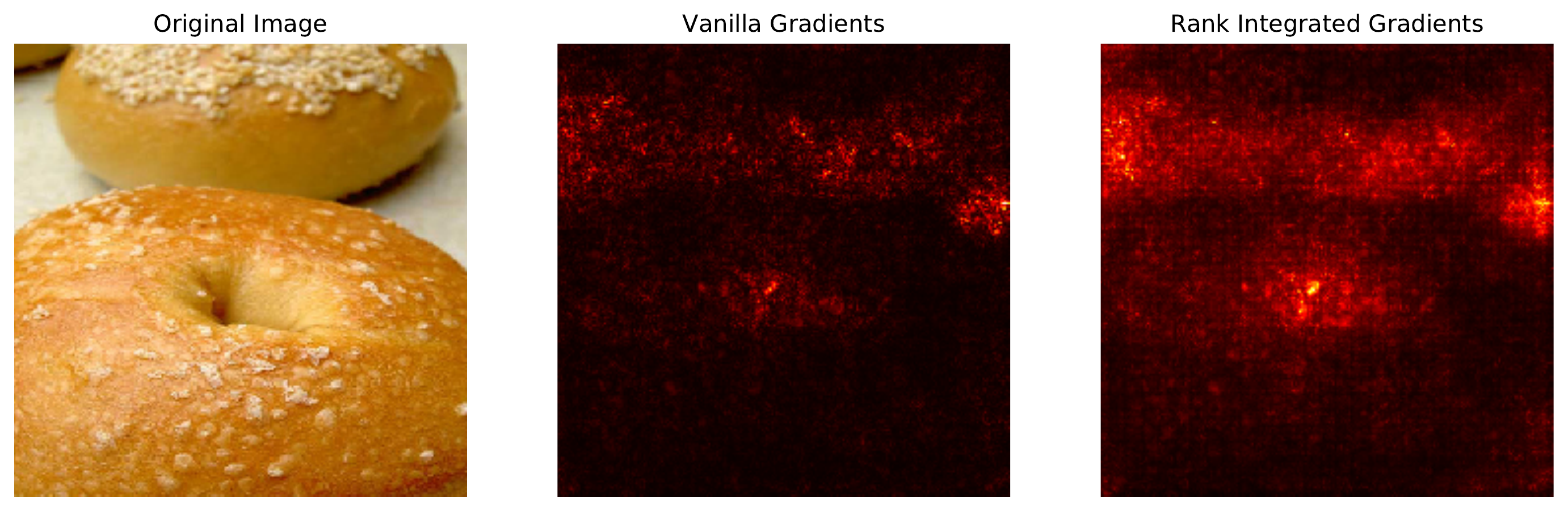}
        \caption{Original label: Bagel}
    \end{subfigure}
    
    \hspace{20pt}
    \begin{subfigure}[b]{0.7\textwidth}
        \centering
        \includegraphics[width=\textwidth]{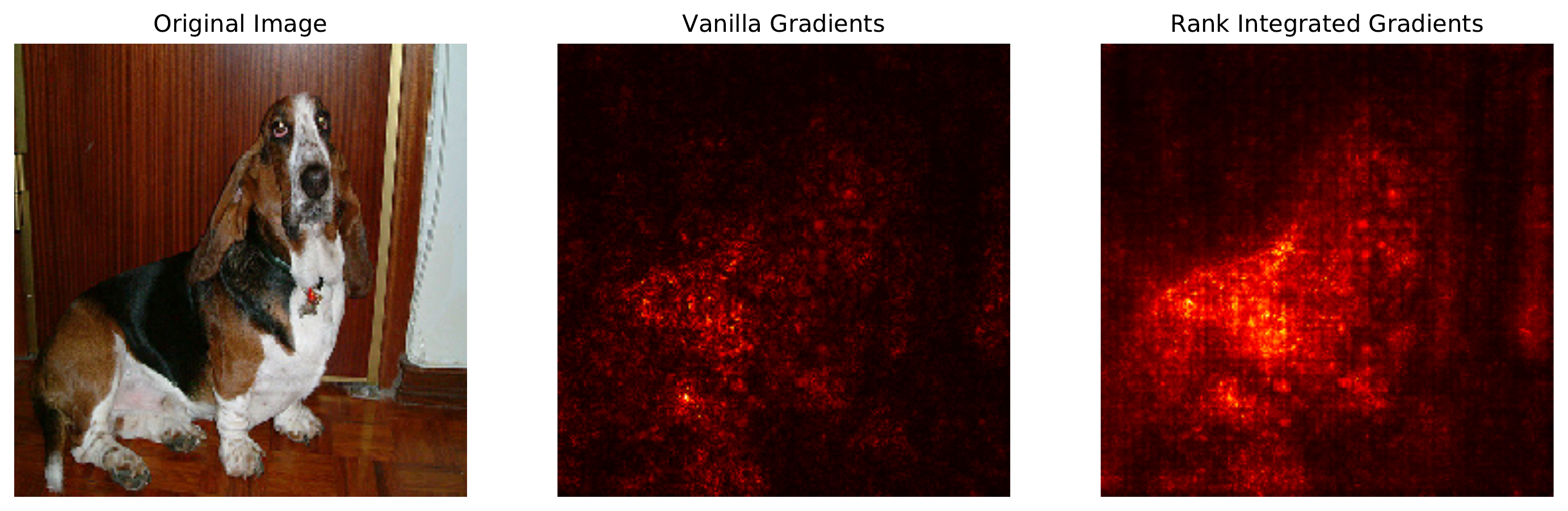}
        \caption{Original label: Basset hound}
    \end{subfigure}
    
    \hspace{20pt}
    \begin{subfigure}[b]{0.7\textwidth}
        \centering
           \includegraphics[width=\textwidth]{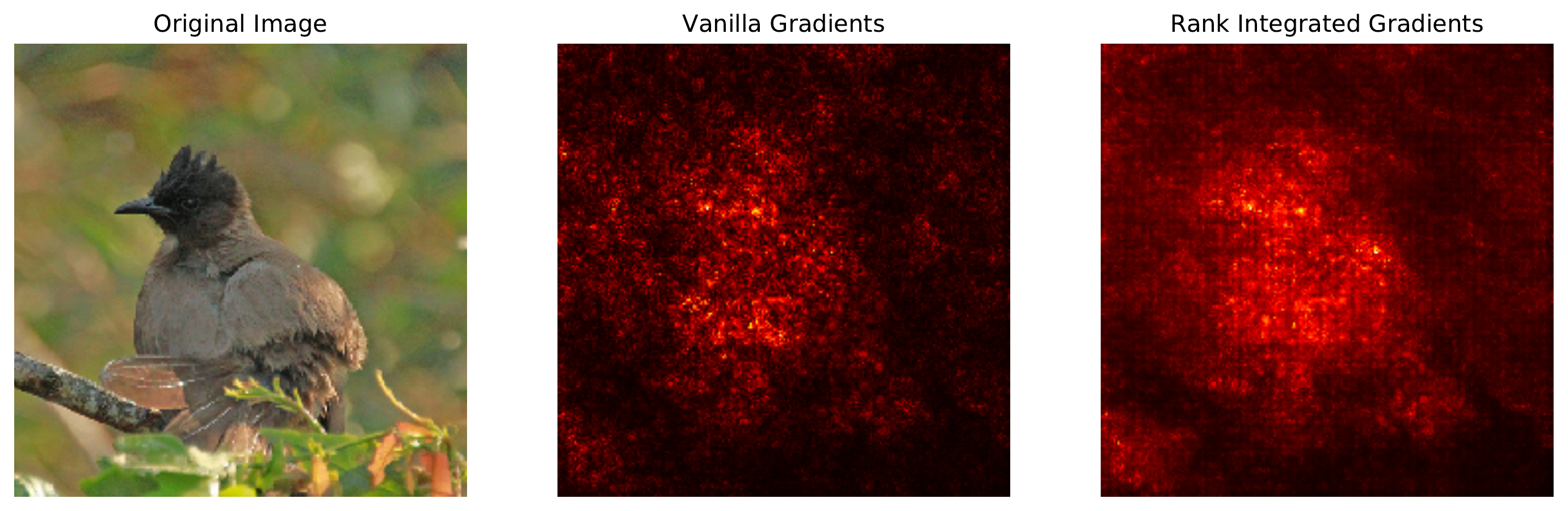}
        \caption{Original label: Bulbul}
    \end{subfigure}
    
    \hspace{20pt}
    \begin{subfigure}[b]{0.7\textwidth}
        \centering
        \includegraphics[width=\textwidth]{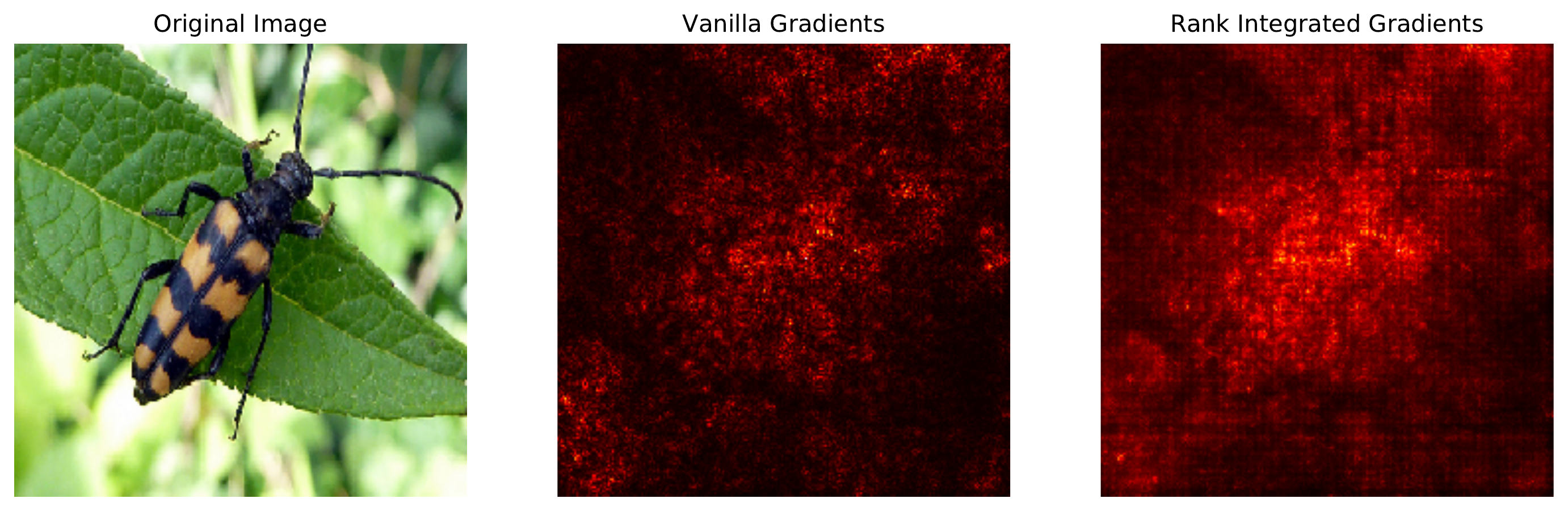}
        \caption{Original label: Long-horned beetle}
    \end{subfigure}
    
    
    \caption{RIG plots for naturally trained ResNet-50 in PyTorch for images randomly chosen from the ImageNet validation set.}
    \label{fig:rig1}
    
\end{figure*}

\begin{figure*}[ht!]
    \centering
    
    \hspace{20pt}
    \begin{subfigure}[b]{0.7\textwidth}
        \centering
        \includegraphics[width=\textwidth]{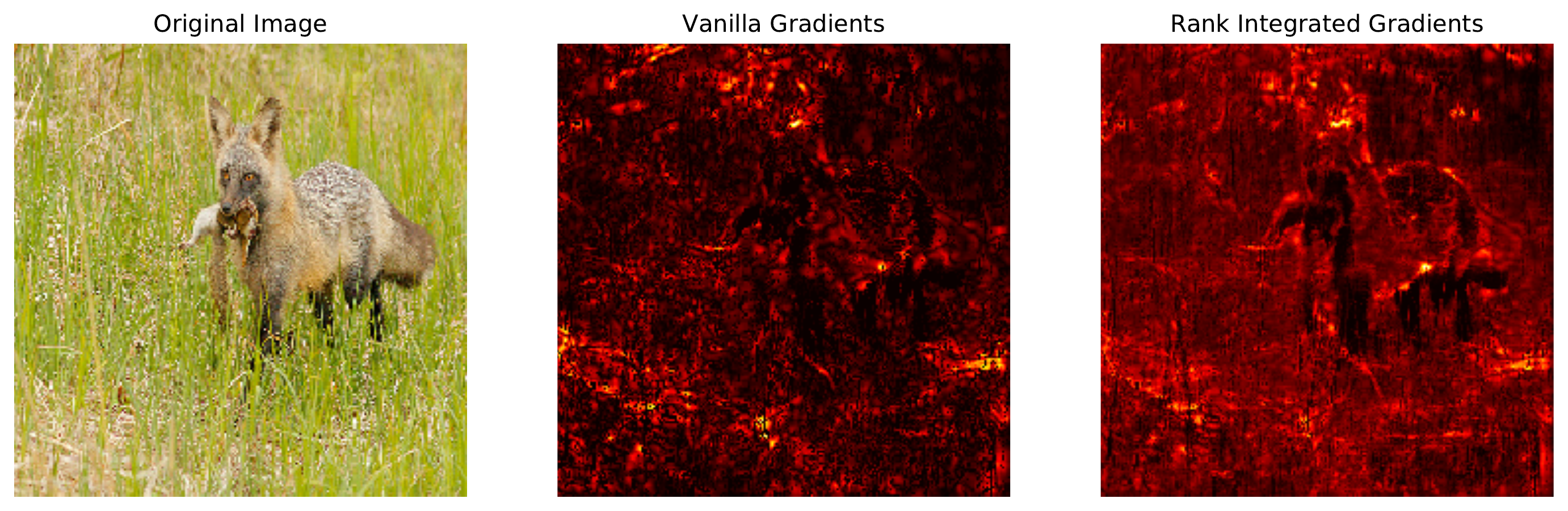}
        \caption{Original label: grey fox}
    \end{subfigure}
    
    \hspace{20pt}
    \begin{subfigure}[b]{0.7\textwidth}
        \centering
        \includegraphics[width=\textwidth]{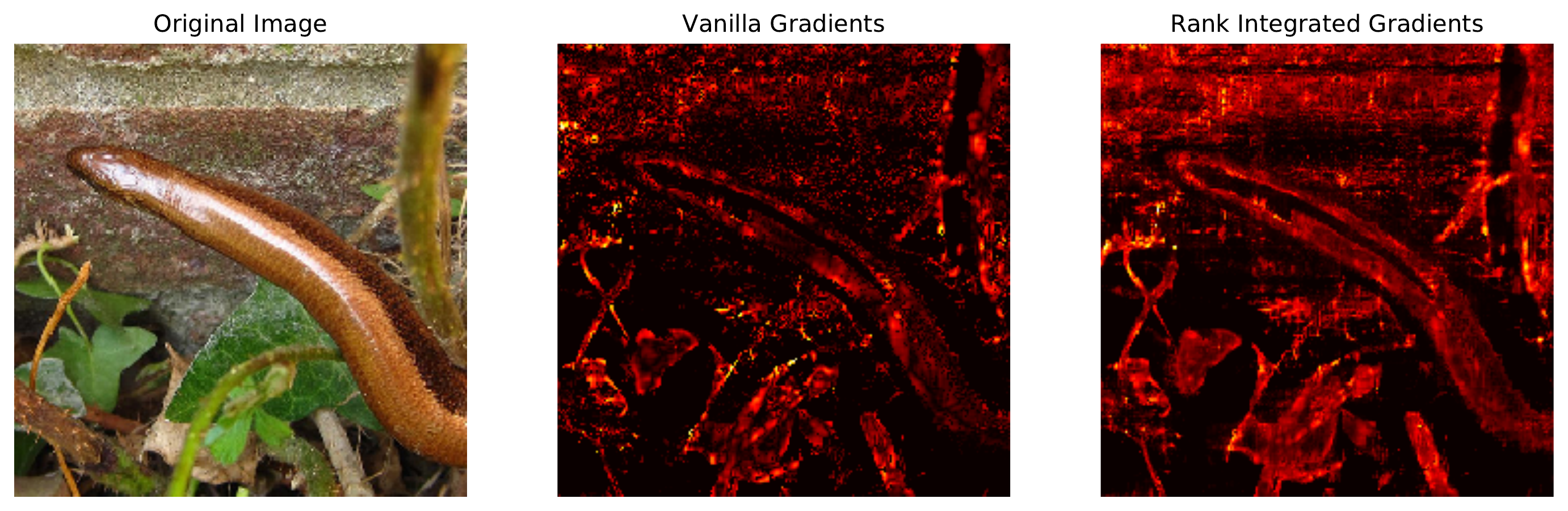}
        \caption{Original label: thunder snake}
    \end{subfigure}
    
    \hspace{20pt}
    \begin{subfigure}[b]{0.7\textwidth}
        \centering
           \includegraphics[width=\textwidth]{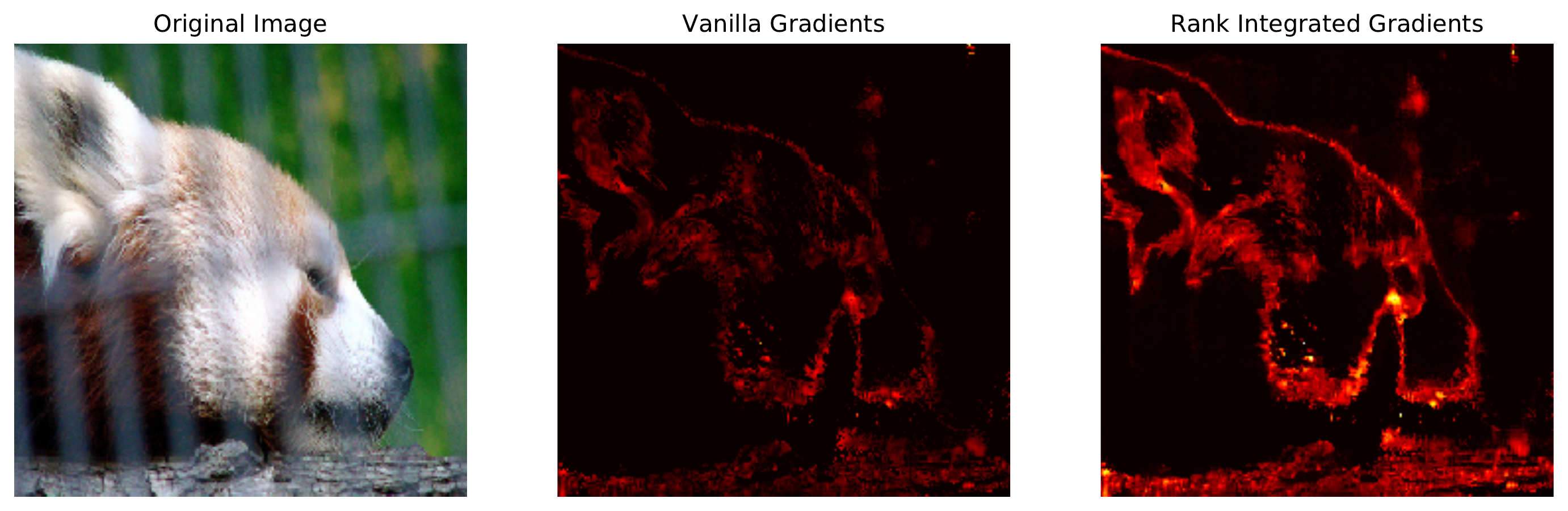}
        \caption{Original label: lesser panda, red panda}
    \end{subfigure}
    
    \hspace{20pt}
    \begin{subfigure}[b]{0.7\textwidth}
        \centering
        \includegraphics[width=\textwidth]{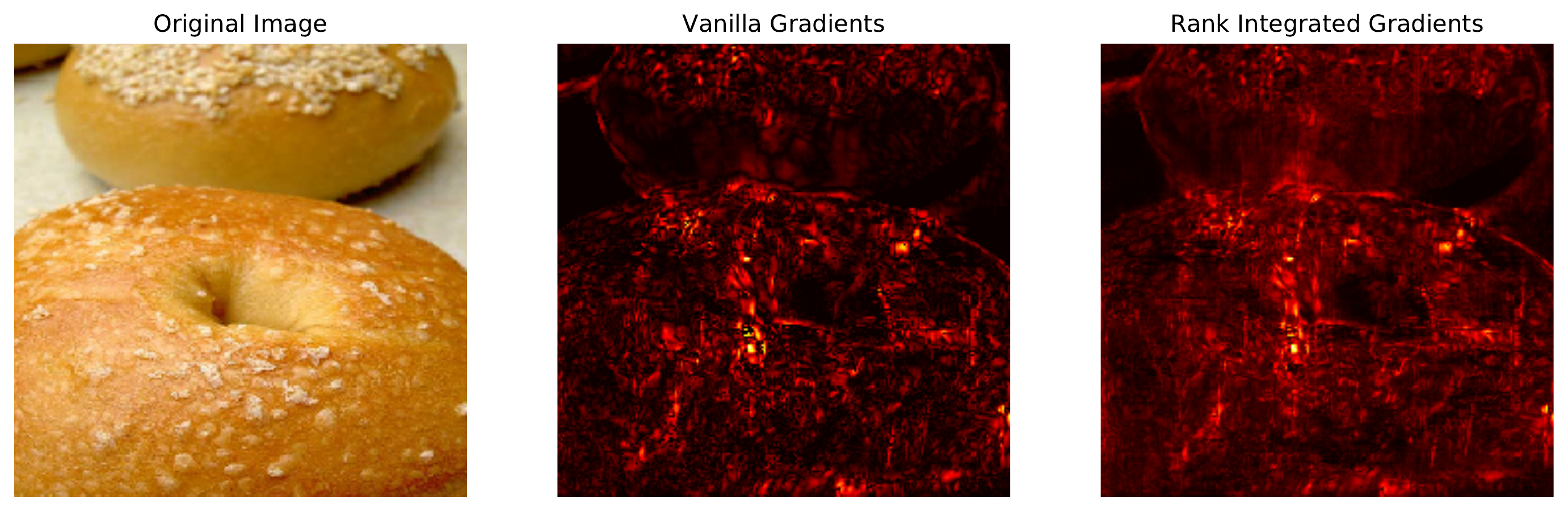}
        \caption{Original label: bagel}
    \end{subfigure}
    
    
    \caption{RIG plots for $L_{2}=3.0$ robust ResNet-50~\cite{robustness} for images randomly chosen from the ImageNet validation set.}
    \label{fig:rig2}
    
\end{figure*}

We provide several more examples of RIG saliency maps for robust and non-robust ResNet-50 models here (Figures~\ref{fig:RIG_EXPLAINED},~\ref{fig:RIG_joint},~\ref{fig:rig1},~\ref{fig:rig2}). RIG highlights rank-based features which are more perceptually-aligned than vanilla gradients for naturally trained as well as adversarially robust networks.

\subsection{Runtime measurements for low-rank approximation}

\begin{figure}[ht!]
    \centering
    \includegraphics[width=0.45\textwidth]{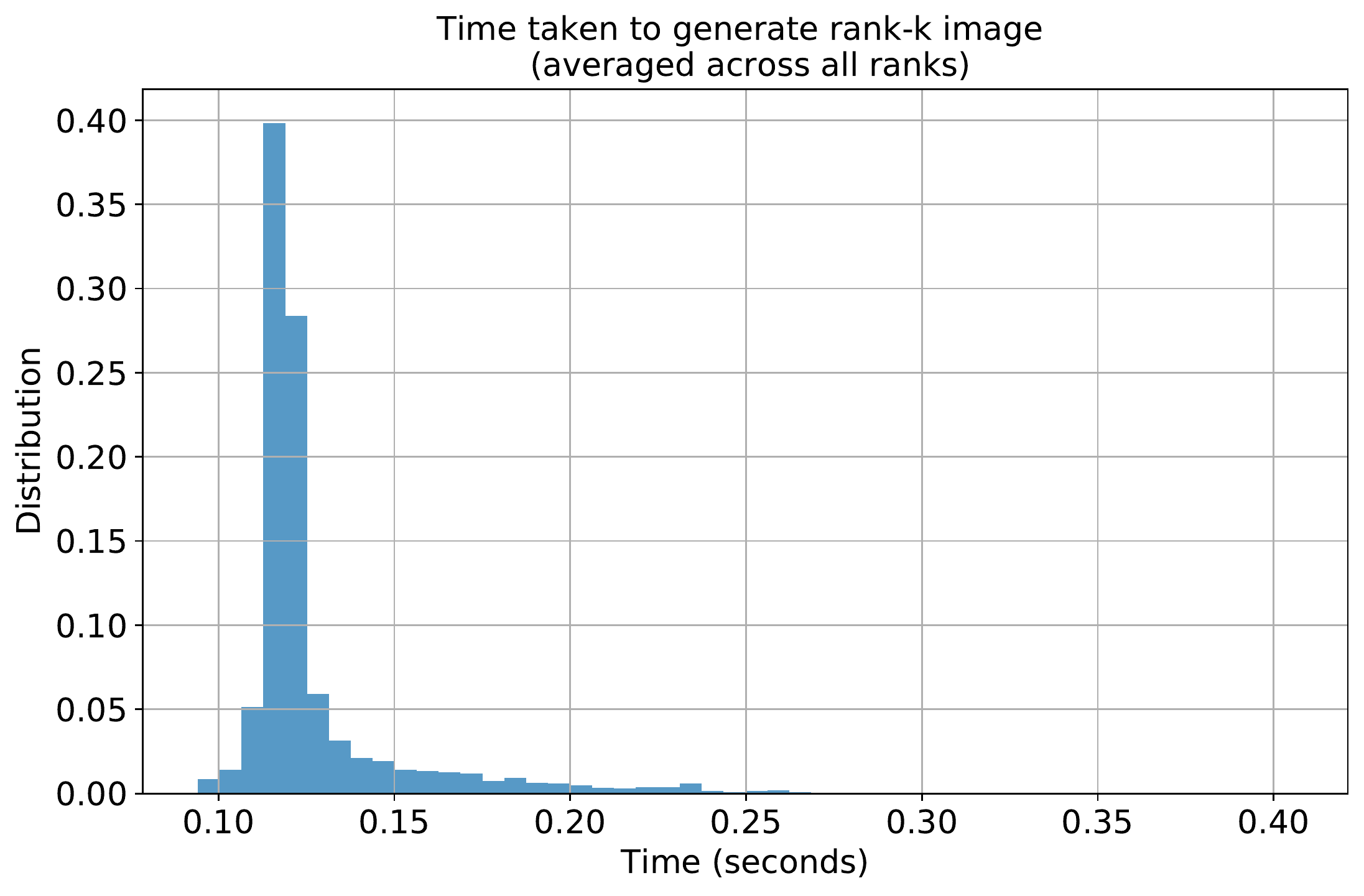}
    \caption{Time required to generate low-rank approximations of $(224 \times 224 \times 3)$ RGB images for ImageNet. Averaged across all possible ranks for $10$ images.}
    \label{fig:time_taken}
\end{figure}

\noindent There is minimal overhead to generating low-rank representations for images, with a distribution over $10$ images across all possible ranks shown in Figure~\ref{fig:time_taken}. The time required to generate rank-$k$ approximations is independent of $k$, and generating arbitrary rank-$k$ representations of a $(224 \times 224 \times 3)$ RGB image for ImageNet inference takes less than $1$ second on an NVIDIA TITAN Xp GPU. 

\begin{table}[ht!]
\centering
\begin{tabular}{c|c}
\hline
\textbf{\begin{tabular}[c]{@{}c@{}}\# of Largest\\ Singular values\\ Truncated?\end{tabular}} & \textbf{\begin{tabular}[c]{@{}c@{}}Test \\ accuracy\end{tabular}} \\ \hline
\textbf{2} & 31.71\% \\ \hline
\textbf{3} & 28.02\% \\ \hline
\textbf{5} & 13.54\% \\ \hline
\textbf{10} & 11.19\% \\ \hline
\end{tabular}
\caption{Test accuracy on full-rank CIFAR-10 test set for ResNet-50 trained on images with largest singular values removed from image.}
\label{tab:truncated_large_vals}
\end{table}

\subsection{Experimental results for VGG-19 and DenseNet}

\begin{figure}[ht!]
    \centering
    \includegraphics[width=0.45\textwidth]{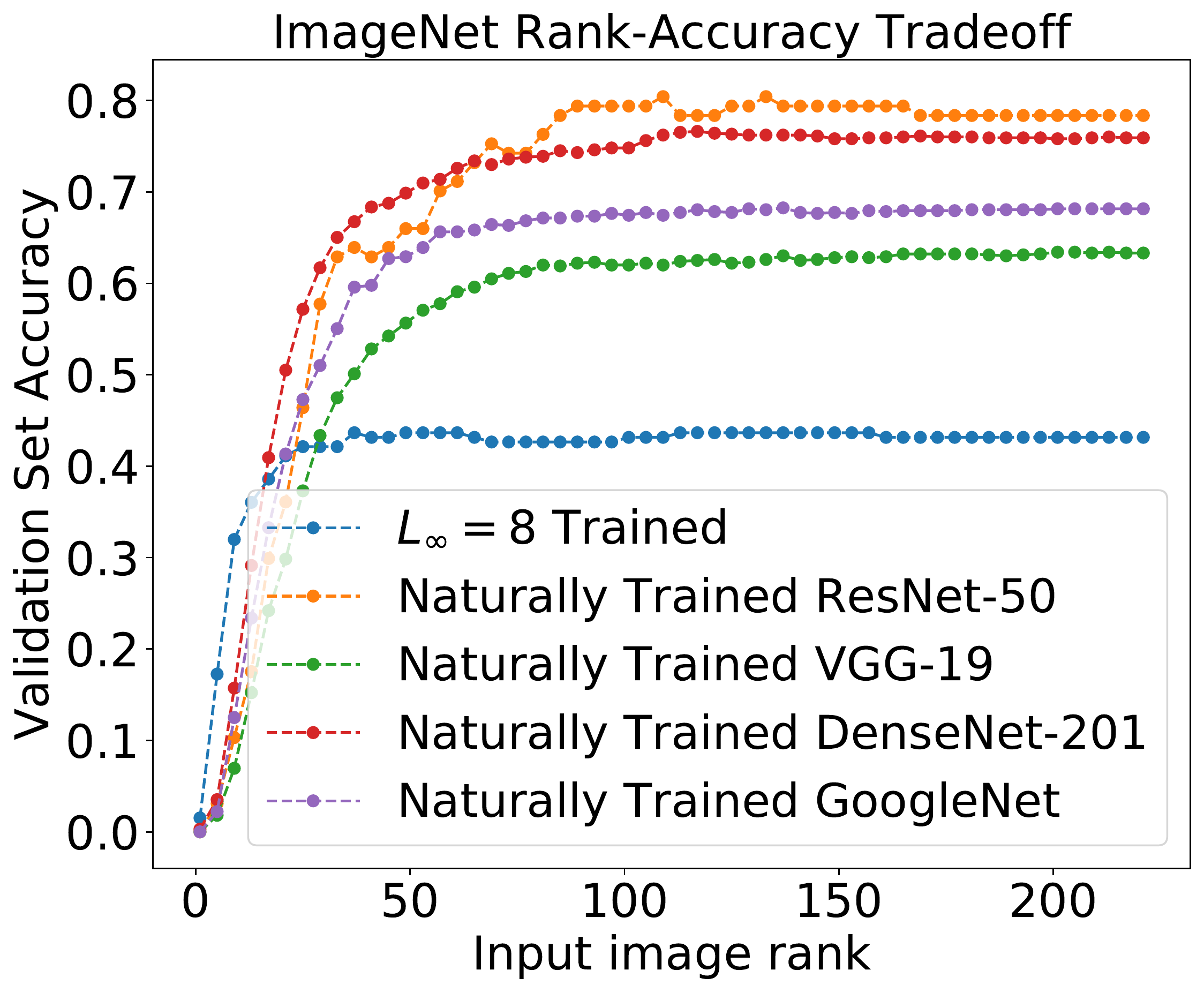}
    \caption{Rank spectrum for naturally trained ResNet-50, DenseNet-201, VGG-19 and GoogleNet architectures compared to a $L_{\infty}=8/255$ robust ResNet-50. }
    \label{fig:multiple_models_rank_spectrum}
\end{figure}

We observe that other state of the art models such as DenseNet~\cite{huang2017densely} and VGG-19~\cite{simonyan2014very} have similar rank-behavior to ResNet-50 models. In particular, we observe in Figure~\ref{fig:multiple_models_rank_spectrum} that VGG-19 is more biased towards higher ranked representations, indicating a potentially larger vulnerability to adversarial examples.

\subsection{Training on solely high-rank representations}
\begin{figure}[h!]
    \centering
    \includegraphics[width=0.45\textwidth]{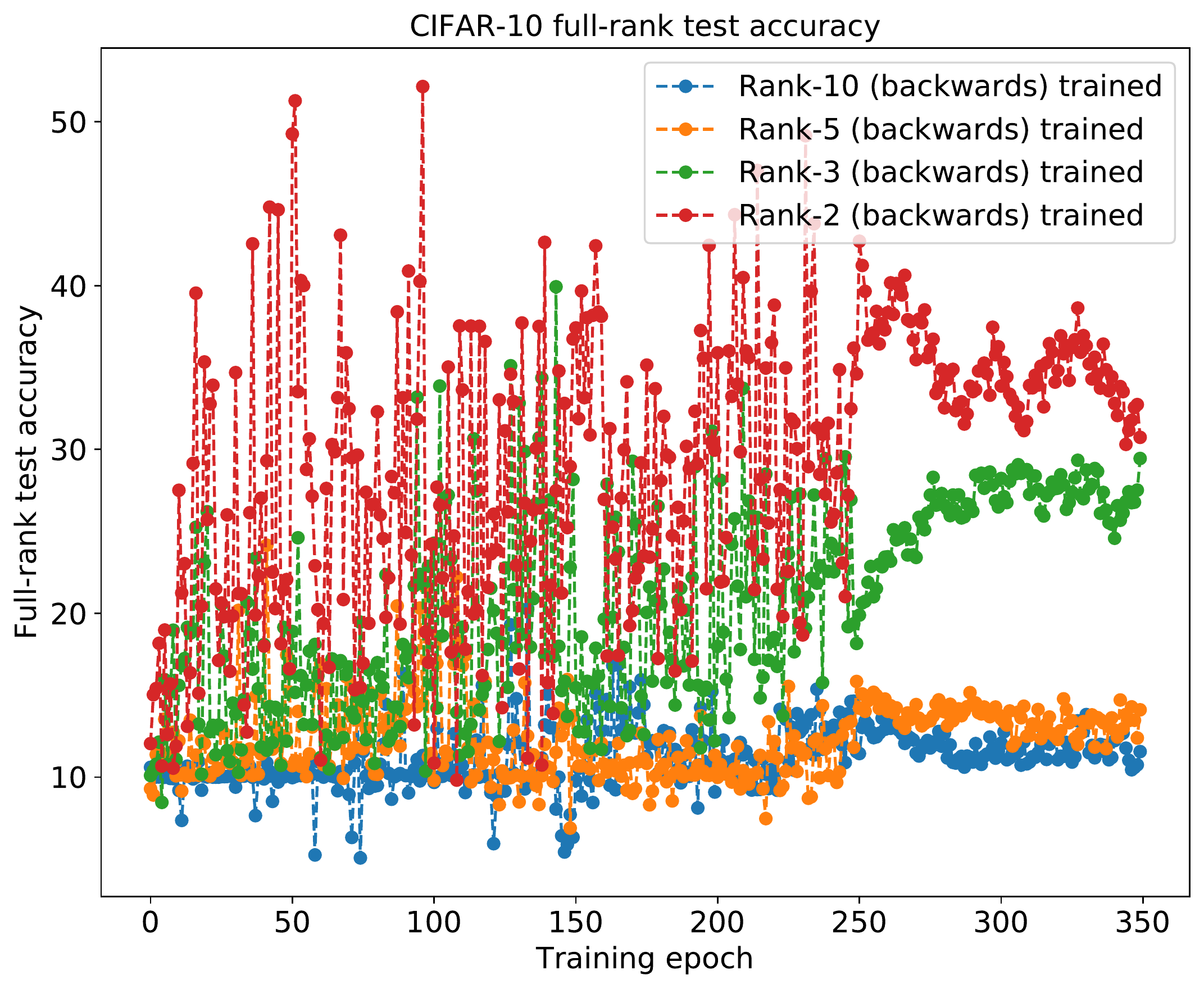}
    \caption{Full-rank test accuracy for ResNet-50 models trained where the largest $10, 5, 3, 2$ singular values from each training image of CIFAR-10 are deleted.}
    \label{fig:high_rank_training}
\end{figure}

Table~\ref{tab:truncated_large_vals} has full-rank test accuracies for ResNet-50 trained on modified versions of the CIFAR-10 dataset, where the largest $k$ singular values for each image are deleted, leaving only higher-ranked features. Test accuracy as a function of training epoch can be found in Figure~\ref{fig:high_rank_training}.

\subsection{Training on solely low-rank representations}
\subsubsection{ImageNet}
\begin{figure}[h!]
    \centering
    \includegraphics[width=0.45\textwidth]{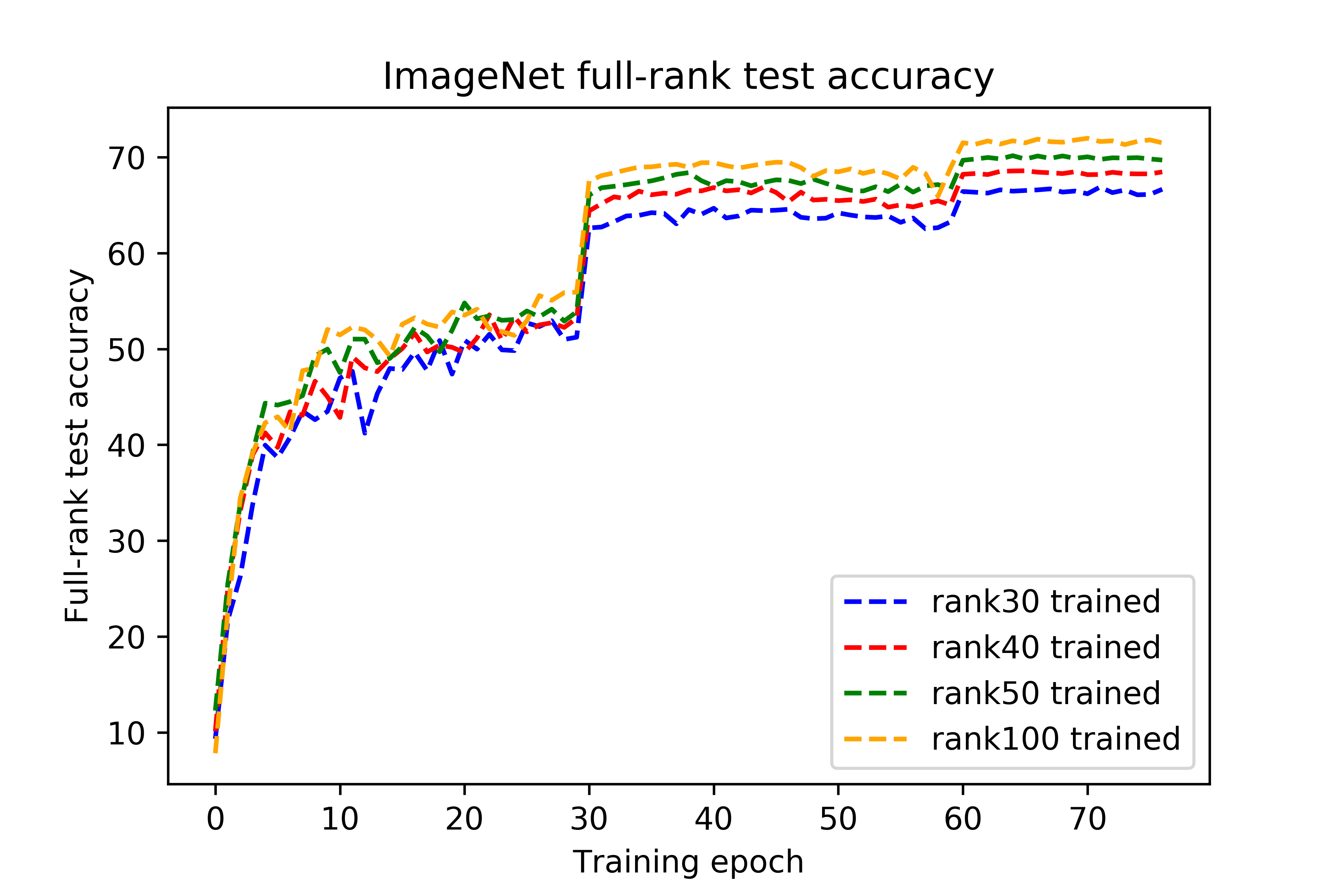}
    \caption{Full-rank test accuracy for ResNet-50 models trained on rank-$30,40,50,100$ ImageNet datasets.}
    \label{fig:low_rank_training_imagenet}
\end{figure}

Figure~\ref{fig:low_rank_training_imagenet} highlights full-rank test accuracy for ResNet-50 on modified low-rank versions of the ImageNet dataset. As expected, the performance of the models increases with increasing ranks of the images. We show this on ranks $30,40,50$ and $100$. For efficient training of ResNet-50 on the various ranks, we pre-process and store the low-rank copies of ImageNet. We trained each network for $24$ hours on $4$ NVIDIA V-$100$ GPUs.

\subsubsection{CIFAR-10}
\begin{figure}[h!]
    \centering
    \includegraphics[width=0.45\textwidth]{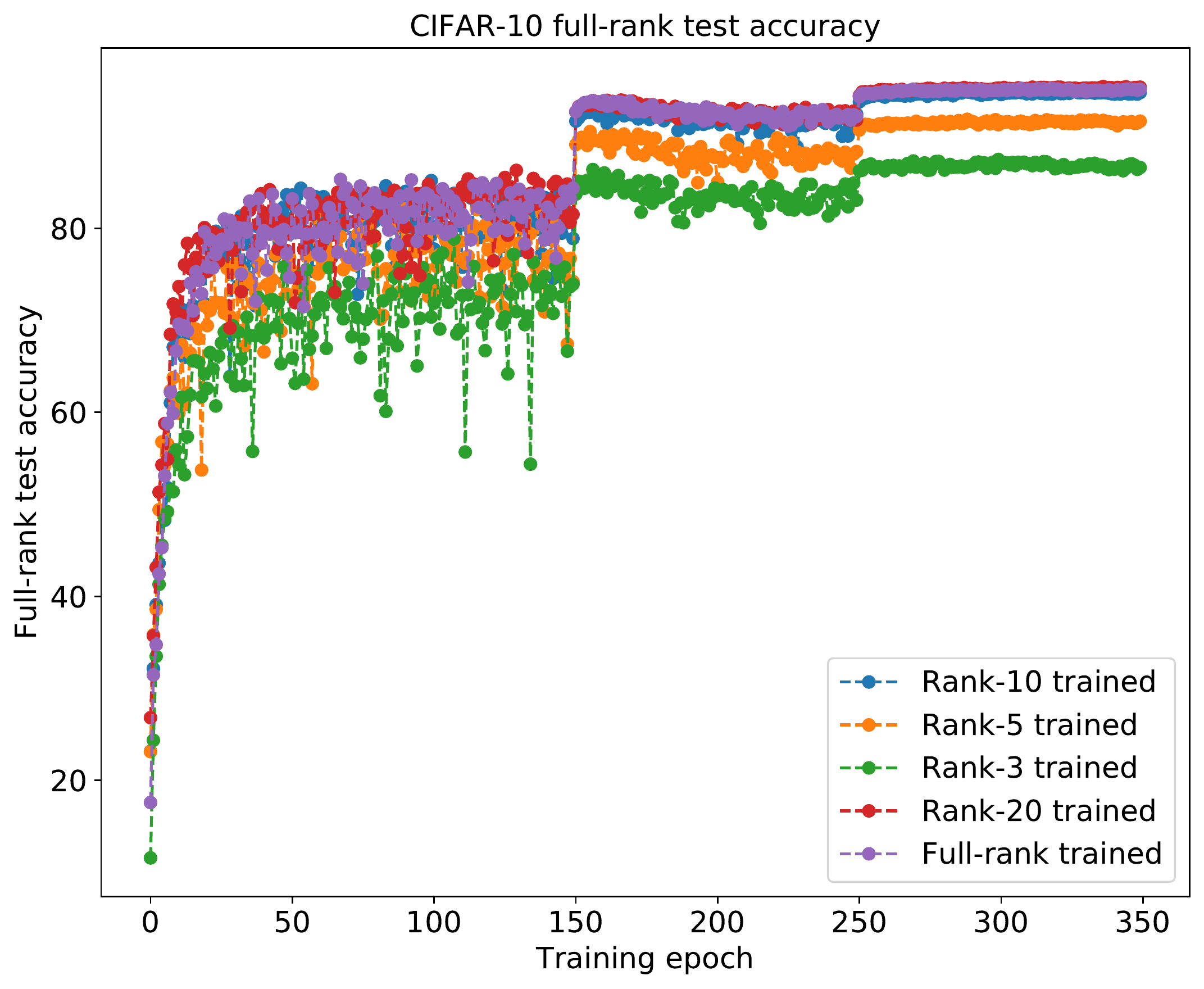}
    \caption{Full-rank test accuracy for ResNet-50 models trained on rank-$3,5,10,20$ CIFAR-10 datasets.}
    \label{fig:low_rank_training}
\end{figure}

Figure~\ref{fig:low_rank_training} highlights full-rank test accuracy for ResNet-50 trained on modified versions of the CIFAR-10 dataset. Models trained on rank-$10,20$ and full rank have identical test accuracies, indicating that a large component of higher-ranked features do not contribute as much to prediction as those from ImageNet.

\end{document}